\documentclass[10pt,twocolumn,letterpaper]{article}

\usepackage{booktabs}
\usepackage{graphicx}
\usepackage{arydshln}
\usepackage{amsmath,amssymb}
\usepackage{caption}
\usepackage{subcaption}
\usepackage{iccv}
\usepackage{dblfloatfix}
\usepackage[numbers,sort&compress]{natbib}

\input{mlpreamble}
\definecolor{iccvblue}{rgb}{0.21,0.49,0.74}
\usepackage[breaklinks,colorlinks,allcolors=iccvblue]{hyperref}

\newcommand{\modelname}{MatPhys}

\title{MatPhys: Learning Material-Aware Physics Parameters for Deformable Object Simulation from Videos}

\author{
Yang Yang\textsuperscript{1,3} \quad
Yiyan Wang\textsuperscript{2,3} \quad
Zheming Liu\textsuperscript{3} \quad
Naoya Iwamoto\textsuperscript{3}\thanks{Corresponding Author} \\
\textsuperscript{1}The University of Osaka \quad 
\textsuperscript{2}The University of Tokyo \quad \\
\textsuperscript{3}Huawei Technologies Japan K.K.
}
\begin{document}

\twocolumn[{
\maketitle
\begin{center}
    \fontsize{9pt}{11pt}\selectfont
    \def\svgwidth{\linewidth}
    \includegraphics[width=\linewidth]{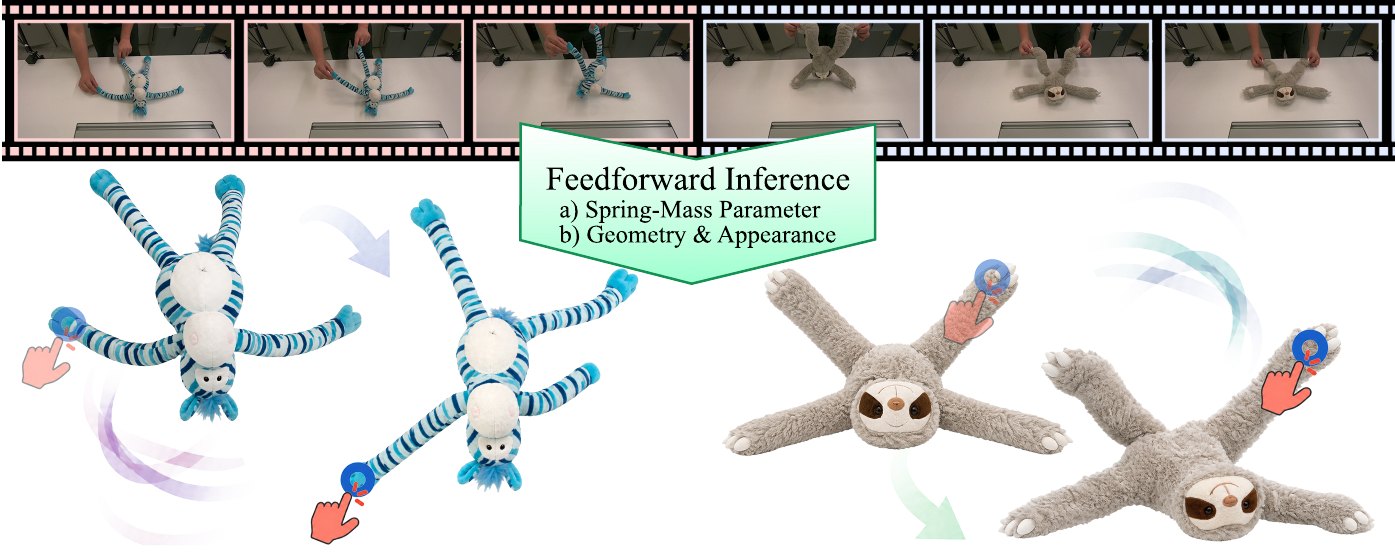}  
    \vspace{-2mm}
    \captionof{figure}{Given a single-view video of a deformable object under interaction, \modelname reconstructs a fully simulatable digital twin with complete geometry, high-fidelity appearance, and accurate physical parameters. Once reconstructed, the digital twin supports intuitive interactive editing: users can freely place control points on the object and manipulate them through drag-based interaction.}  
    \label{fig:teaser}
\end{center}
}]

\begin{abstract}
Reconstructing simulation-ready deformable objects is important for vision, graphics, and robotics. Existing physics-driven methods can recover physical digital twins from videos, but they suffer from two fundamental limitations: they typically assume a homogeneous material across the whole object, and their scene-specific inverse optimization, combined with the inherent ambiguity of monocular observation, yields inconsistent parameters for the same material across different scenes or interactions. We propose \modelname, a material-aware feed-forward framework that predicts spring-mass parameters from a single-view video, addressing these two issues with two coupled designs. To relax the homogeneous material assumption, we use DINO features to decompose the object into semantically meaningful parts and to query a part-level material prior, assigning each part its own physical behavior. To enforce cross-scene consistency, we introduce a learned material codebook of shared material embeddings as the bridge between appearance and physics, and further use the part-level prior as a reference distribution that constrains the decoder so that the same material yields consistent parameters across scenes and interactions. Together, these designs turn an under-constrained monocular problem into feed-forward inference grounded on shared, reusable material concepts.
Experiments show that our method matches per-scene optimization baselines in reconstruction and future prediction, while achieving stronger generalization to unseen interactions and objects with more consistent physical parameters.
\end{abstract}
\vspace{-0.8em}
\section{Introduction}
Creating interactive physical digital twins from visual observations is an important problem in computer vision, graphics, and robotics. A 3D asset for interaction and simulation should not only have high-quality geometry and appearance, but also physical properties that allow it to respond to external forces and manipulation. Such assets can support future prediction, interactive editing, and large-scale simulation data generation. However, recovering simulation-ready 3D assets from limited visual input remains challenging.

Recent progress in NeRF~\cite{mildenhall2020nerf, Gao-ICCV-DynNeRF, pumarola2020d,Zhang2025RealtoSimRP}, 3D Gaussian Splatting~\cite{kerbl3Dgaussians}, and their dynamic extensions~\cite{zhao2024physplat,luiten2023dynamic, 4dgs,park2021nerfies,park2021hypernerf, tretschk2021nonrigid,yang2024deformable3dgs,huang2024scgs} has greatly improved dynamic scene reconstruction. These methods can recover high-quality geometry, appearance, and observed motion from images or videos. However, they usually model motion as visual deformation or temporal change, without explicitly recovering the physical parameters behind the object. To bridge rendering and simulation, recent works introduce physical models into 3D representations. Some learning-based methods~\cite{xie2024physgaussian,gsdynamics,4dgs, huang2024scgs} use neural networks to model object dynamics directly. They have strong fitting ability, but often require large amounts of training data and are limited to specific objects and actions. 

Other physics-driven methods couple 3D representations with explicit simulators and estimate physical parameters through differentiable inverse optimization. 
MPM-based methods~\cite{stomakhin2013mpm, liu2024physics3d, chenhu2026empm} provide expressive continuum dynamics for complex deformation, while spring-mass systems~\cite{terzopoulos1987elastically, baraff1998large, zhong2024springgaus} offer a lightweight and interpretable alternative with explicit stiffness, damping, topology, and contact parameters. 
Recent spring-mass Gaussian methods, including Spring-Gaus~\cite{zhong2024springgaus}, PhysTwin~\cite{jiang2025phystwin}, and NeuSpring~\cite{xu2026neuspring}, demonstrate simulation-ready reconstruction from videos by combining 3D Gaussians with inverse physical fitting. 
However, their parameters are still optimized in a scene-specific manner, which can lead to inconsistent estimates for similar materials across scenes and interactions. 
Moreover, geometry-driven decomposition, such as KNN-based clustering, does not necessarily align physical regions with semantic material parts.

This raises a natural question: can deformable-object physics be inferred in a feed-forward manner from monocular RGB video? Recent concurrent works suggest that this direction is promising. ReconPhys~\cite{wang2026reconphys} shows that learned predictors can transfer physical estimates across objects, while PhysSplat~\cite{zhao2024physplat} shows that MLLMs provide useful visual priors about material and physical properties. However, neither work learns an explicit, transferable material representation for predicting simulator parameters: ReconPhys relies on global video conditioning in synthetic gravity-driven settings, and PhysSplat queries object-level material priors that collapse heterogeneous parts into a single material assumption. These limitations leave open two key challenges: identifying semantic parts that correspond to changes in different physical behavior, and maintaining consistent physical parameters for the same material across scenes and interactions.

To address these two issues, we propose \modelname, a material-aware feed-forward framework that predicts spring-mass simulation parameters from monocular RGB video. Our key idea is to use semantic parts as the unit for material reasoning: instead of assigning one material prior to the whole object, each part receives its own material evidence and can induce different local spring behavior, which relaxes the homogeneous material assumption and provides a structured interface between visual appearance and physical prediction.
Concretely, given an input video, we first reconstruct a 3D Gaussian object representation from a keyframe using a feed-forward 3D reconstruction model~\cite{xiang2025trellis2}, and use the Gaussians as the geometry representation for simulation. We then lift DINO features~\cite{Oquab2023DINOv2LR} onto the object and cluster them to obtain semantically meaningful parts. For each part, we query an MLLM to obtain a material class prior and a distribution of its corresponding physics parameters. We further estimate a part-aware topology and abstract the Gaussians as mass points, yielding a spring-mass system.

On top of this representation, we train a feed-forward physics predictor to estimate the spring-mass parameters. We argue that the prediction should be conditioned on three complementary signals: material features, which capture transferable physical tendencies; motion features, which reveal how the object responds under interaction; and local geometric features, which encode the edge scale, orientation, and neighborhood structure that modulate local spring behavior. To obtain transferable material representations, we introduce a learnable material codebook shared by all objects and scenes. Each part-level material class prior is mapped through this codebook into a material embedding, which serves as the material representation. In parallel, a pretrained video encoder~\cite{tong2022videomae} along with a small projector extracts motion features from the input video, and a geometry encoder computes local edge features from the mass-point coordinates and spring connectivity. A physics decoder fuses these features to predict per-edge stiffness, controller-object stiffness, and global damping/contact parameters. During training, we minimize the discrepancy between simulated results and observations while using the part-level physical distribution as a reference for constraining the physical consistency.

Together, these designs enable \modelname to learn transferable material-to-physics relations and predict simulation-ready parameters in a single forward pass. Experiments show that our method matches per-scene optimization baselines in reconstruction and future prediction, while achieving stronger generalization to unseen objects.

Our contributions are threefold: 
\begin{itemize}
\item We propose \modelname, a material-aware feed-forward framework for predicting simulation-ready digital twins from monocular RGB video.
\item We introduce a learnable material codebook, enabling transferable material-to-physics prediction and more consistent physical parameters across scenes and interactions.
\item We use semantic part decomposition and part-level MLLM material queries to build part-aware spring-mass models, relaxing the homogeneous material assumption for heterogeneous deformable objects.
\end{itemize}

\section{Related Work}
\subsection{Physics-based simulation of deformable objects}
Physics-based deformable modeling has long relied on explicit simulators such as spring-mass systems and MPM. Recent vision-driven work has integrated such simulators with 3D reconstruction to recover deformable digital twins from videos. Spring-Gaus~\cite{zhong2024springgaus} first couples spring-mass dynamics with 3D Gaussians to reconstruct and simulate elastic objects. PhysTwin~\cite{jiang2025phystwin} extends this line to robotics-oriented digital twins, combining spring-mass simulation, complete-geometry recovery, and rendering from sparse-view interaction videos. NeuSpring~\cite{xu2026neuspring} further introduces piecewise spring topology and neural spring fields for heterogeneous deformable objects. These methods demonstrate that spring-mass simulation is practical for reconstruction and interaction, but they remain largely scene-specific and rely on staged or heuristic topology construction. On the continuum side, Physics3D~\cite{liu2024physics3d}, DreamPhysics~\cite{Huang2024DreamPhysicsLP}, and EMP\cite{chenhu2026empm} show that MPM-based models can capture richer material behavior and larger deformation. However, their focus is primarily on physical realism, scene-specific fitting, or stronger observation settings, rather than transferable semantic-material-conditioned parameter prediction.

\subsection{Learning-based deformation modeling}
Another line of work learns deformable dynamics from data instead of optimizing physical parameters for each sequence.
PhysWorld~\cite{Yang2025PhysWorldFR} builds an MPM-based digital twin from videos and augments it with physically plausible demonstrations to train a lightweight GNN world model for fast prediction.
Latent Intuitive Physics~\cite{Zhu2024LatentIP} learns hidden physical properties of fluids as latent variables and transfers them to novel simulation scenarios.
While these methods show promising transferable dynamics, the physical properties are often embedded in neural predictors or implicit latent spaces.
In contrast, we predict explicit part-level, material-aware parameters for a spring-mass Gaussian representation, enabling interpretable and simulation-ready reconstruction from single-view videos.

\subsection{Lifting 2D semantic features into 3D}
Recent methods lift semantic priors from 2D foundation models into 3D representations for open-vocabulary understanding and part-level grouping.
LERF~\cite{kerr2023lerf} embeds CLIP features into NeRF for language-based 3D queries, while GARField~\cite{kim2024garfield} learns 3D affinity fields from SAM masks for hierarchical grouping.
For Gaussian representations, Feature 3DGS~\cite{zhou2024feature3dgs} and LangSplat~\cite{qin2024langsplat} distill 2D visual-language features into 3D Gaussians, enabling semantic querying and editing.
Gaussian Grouping~\cite{ye2024gaussian} and SAGA~\cite{cen2025saga} further use 2D masks or prompts for object-level Gaussian segmentation.
Our work follows this direction by lifting DINO features into 3D Gaussians, but uses them as conditioning cues for material-aware physical parameter prediction rather than only semantic recognition or segmentation.
\begin{figure*}[t!]
  \centering
  \includegraphics[width=\linewidth]{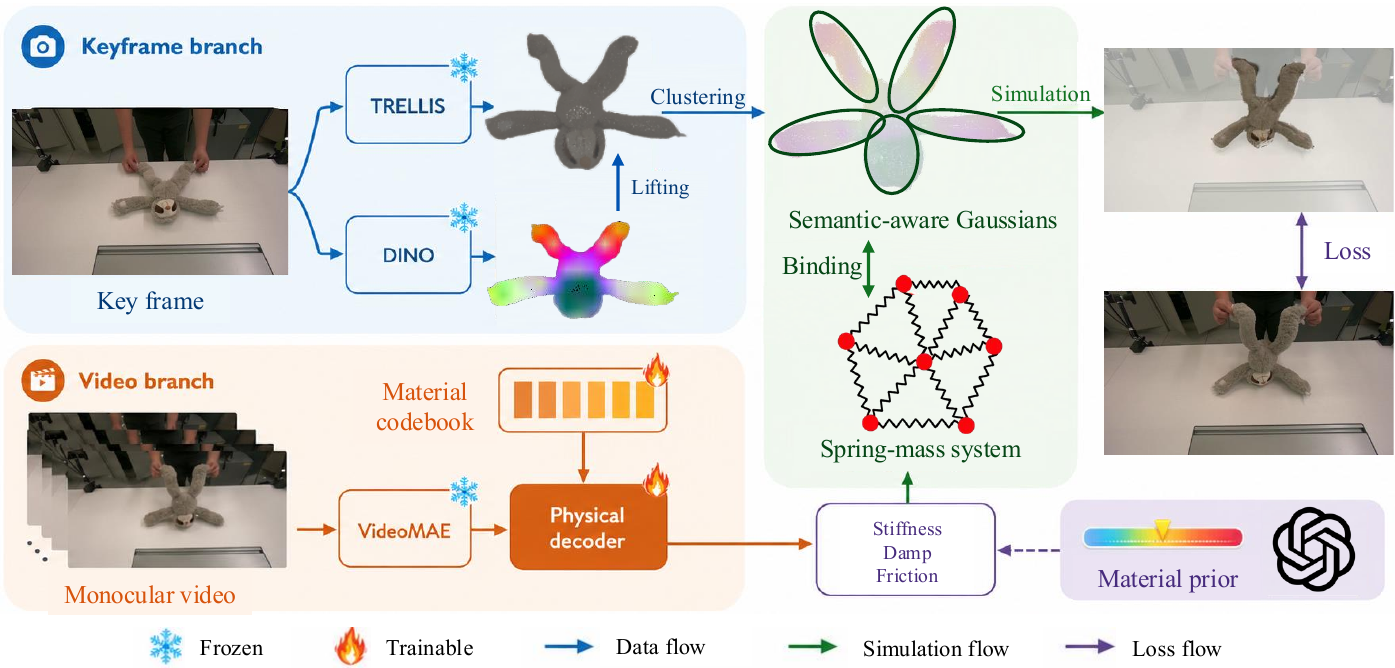}
  \caption{\textbf{Overview of our framework.} Given a monocular video of a deformable object, we use the key frame to reconstruct an explicit 3D Gaussian representation with TRELLIS2 and lift dense DINO features onto the Gaussians for semantic-aware part decomposition. The clustered semantic parts are used to build a part-aware spring-mass topology, where Gaussian centers serve as mass points and spring connections are constructed within and across parts. In parallel, the monocular video is encoded by a video motion encoder, and an MLLM provides part-level material priors, which are embedded through a learnable material codebook. A physical decoder then predicts simulation parameters, including stiffness, damping, and friction, to instantiate the spring-mass simulator. The resulting simulation-ready Gaussian twin can be rolled out to reproduce and predict object deformation under interaction without per-scene inverse optimization.}
  \label{fig:overview}
\end{figure*}
\section{Preliminary: spring-mass model}
A spring-mass model is a typical framework for physically-based dynamic modeling. It represents a deformable object using a system of springs and masses, which can be formulated as a graph $G=(\mathcal{V},\mathcal{E})$, where $\mathcal{V}$ denotes the set of mass points representing object geometry and $\mathcal{E}$ denotes the set of springs defining the object topology. Typically, $\mathcal{E}$ is initialized as a KNN graph over mass-point coordinates and then kept fixed during optimization. Each mass point has a position $\V{x}_i \in \mathbb{R}^3$ and a velocity $\V{v}_i \in \mathbb{R}^3$ describing its motion over time, while each spring connects two mass points with physical parameters such as rest length, stiffness, and damping, where rest length defines the undeformed shape, stiffness controls resistance to stretching or compression, and damping dissipates energy to reduce oscillation. This compact representation provides an efficient basis for simulating deformable object dynamics.

Given the spring graph, the motion of each mass point is governed by Newtonian dynamics. For each mass point $i$, the total force is computed from the springs connected to it, damping forces, and external forces as
  \begin{equation}
  \mathbf{F}_i =
  \sum_{(i,j)\in \mathcal{E}}
  \left(
  \mathbf{F}_{ij}^{\text{spring}} + \mathbf{F}_{ij}^{\text{damping}}
  \right)
  + \mathbf{F}_i^{ext},
  \end{equation}
where the spring force between point $i$, $j$ is given by 
  \begin{equation}
  \mathbf{F}_{ij}^{\text{spring}} =k_{ij}
  \left(
  \|\mathbf{x}_j - \mathbf{x}_i\| - l_{ij}
  \right)
  \frac{\mathbf{x}_j - \mathbf{x}_i}
  {\|\mathbf{x}_j - \mathbf{x}_i\|},
  \end{equation}
  where $k_{ij}$ is the spring stiffness and $l_{ij}$ is the rest length, and the dashpot damping force is given by
  \begin{equation}
  \mathbf{F}_{ij}^{\text{damping}} = -\gamma (\V{v}_i-\V{v}_j),
  \end{equation}
  where $\gamma$ is the dashpot damping coefficient. Then the system updates positions and velocities using explicit Euler integration as
  \begin{equation}
  \mathbf{v}_i^{t+1}
  = \delta(\mathbf{v}_i^t + \Delta t \frac{\mathbf{F}_i}{m_i}),
  \qquad
  \mathbf{x}_i^{t+1}
  = \mathbf{x}_i^t + \Delta t \mathbf{v}_i^{t+1},
  \end{equation}
  where $\delta$ denotes the drag damping.
  In our method, the key physical quantities, including spring stiffness, damping, contact parameters, and topology, are predicted from visual semantics and
  material priors rather than optimized independently for each scene.
\section{Method}

We aim to build a simulation-ready physical twin of a deformable object from monocular RGB video. 
Given an input video, our framework reconstructs the object as a set of 3D Gaussians and predicts its physical parameters in a feed-forward manner.
An overview of the proposed pipeline is shown in Fig.~\ref{fig:overview}. We first represent the object as
\begin{equation}
\mathcal{G} = \{ \mathbf{x}_i, \mathbf{r}_i, \mathbf{s}_i, \mathbf{c}_i, \alpha_i \}_{i=1}^{N},
\end{equation}
where each Gaussian encodes its center, rotation, scale, color, and opacity.
To make this representation simulatable, we treat Gaussian centers as mass points and connect them with springs, forming a physical graph
$G=(\mathcal{V},\mathcal{E})$.
Given the predicted physical parameters $\Theta$, a spring-mass simulator $\mathcal{S}$ rolls out the Gaussian centers over time:
\begin{equation}
\{\hat{\mathbf{x}}_i^t\}_{i=1}^{N} =
\mathcal{S}(\mathcal{G}, \mathcal{E}, \Theta, t),
\end{equation}
where $\Theta$ includes stiffness, damping, and contact-related parameters.
\subsection{Object reconstruction with part decomposition}
Recent feed-forward 3D reconstruction models can recover reasonable object geometry from a single image by leveraging strong learned shape priors. Given a monocular input video, we use the first frame as the reconstruction keyframe and reconstruct the object as 3D Gaussians using TRELLIS2~\cite{xiang2025trellis2} which produces an explicit 3D Gaussian representation in a single forward pass.

However, geometry alone is not sufficient for material-aware physical modeling. Traditional spring-mass constructions often build topology from geometric proximity, such as KNN graphs, or assign physical parameters at the whole-object level. Such geometry-driven representations do not explicitly identify semantic material regions, making it difficult to model heterogeneous objects whose different parts exhibit different physical behavior. We therefore augment the reconstructed Gaussians with semantic features and decompose them into part-level units before querying material priors and predicting physical parameters.

Specifically, we extract dense DINO features from the reconstruction keyframe and lift them onto the reconstructed Gaussians. For each visible Gaussian, we project its center $\mathbf{x}_i$ onto the image and sample the corresponding 2D feature. For invisible Gaussians, we complete their features by symmetry-based propagation when a symmetric counterpart is available. This augments each Gaussian with a semantic feature $\mathbf{f}_i^{\mathrm{sem}}$ while keeping its original geometric and appearance attributes. We then cluster the lifted feature set $\{\mathbf{f}_i^{\mathrm{sem}}\}_{i=1}^{N}$ into $K$ semantic parts:
\begin{equation}
\mathcal{P} = \{P_k\}_{k=1}^{K},
\qquad
P_k = \{ i \mid c_i = k \},
\end{equation}
where $c_i$ is the cluster assignment of Gaussian $i$, and $K$ is a hyper-parameter for determining cluster number. These parts group Gaussians with similar semantic cues and serve as the units for material querying, topology construction, and physical parameter prediction.

\subsection{Material prior query and part-level topology}
Semantic parts are the units at which material identity and local deformation behavior vary. We therefore query part-level material priors and construct topology.
For each part $P_m$, we render a highlighted overlay on the input image and query an off-the-shelf MLLM to obtain a material class distribution over plausible materials, together with a coarse reference distribution of physical properties such as stiffness, damping, friction, and collision elasticity. These queried values are not used as simulator parameters directly, but serve as soft guidance for learning material-to-physics prediction.

We then construct spring connections conditioned on each part. Specifically, for part $P_m$, we estimate topology hyperparameters including the neighborhood size $k_m^{\mathrm{nn}}$ and connection radius $r_m$ from its material prior and local Gaussian density. Softer or denser parts use denser connectivity, while harder or sparser parts use simpler connectivity. 
We build intra-part springs using the corresponding local rule and keep a small number of boundary springs between adjacent parts for global connectivity.
\begin{figure*}[t!]
  \setlength{\tabcolsep}{1pt}
  \renewcommand{\arraystretch}{0}
  \newlength{\qcellw}\setlength{\qcellw}{0.235\textwidth}
  \newcommand{\qimg}[1]{\includegraphics[width=\qcellw]{#1}}
  \newcommand{\qrow}[1]{\rotatebox{90}{\normalsize #1}}

  \begin{tabular}{c@{\hskip 2pt}cc;{1pt/1pt}cc}
    & \multicolumn{2}{c}{\textcolor{blue!80!black}{\normalsize\itshape Reconstruction \& Resimulation} $\rightarrow$}
    & \multicolumn{2}{c}{\textcolor{red!70!black}{\normalsize\itshape Future Prediction} $\rightarrow$} \\[4pt]

    \qrow{Observation}       & \qimg{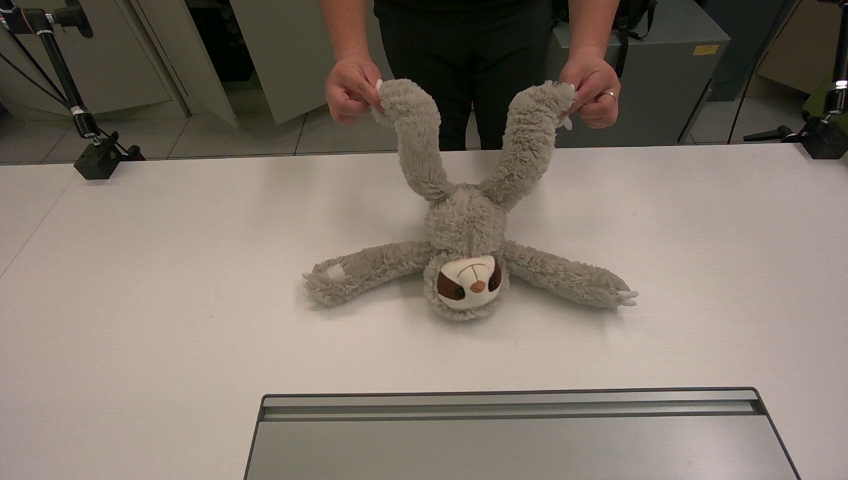}       & \qimg{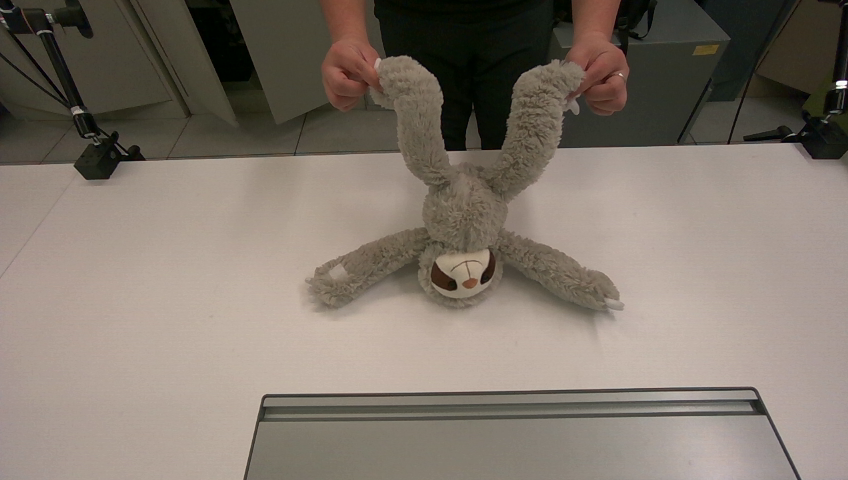}
                             & \qimg{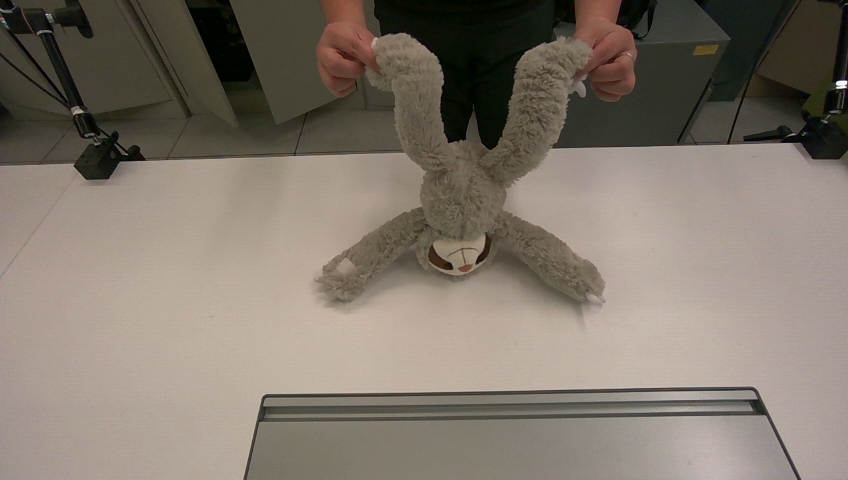}       & \qimg{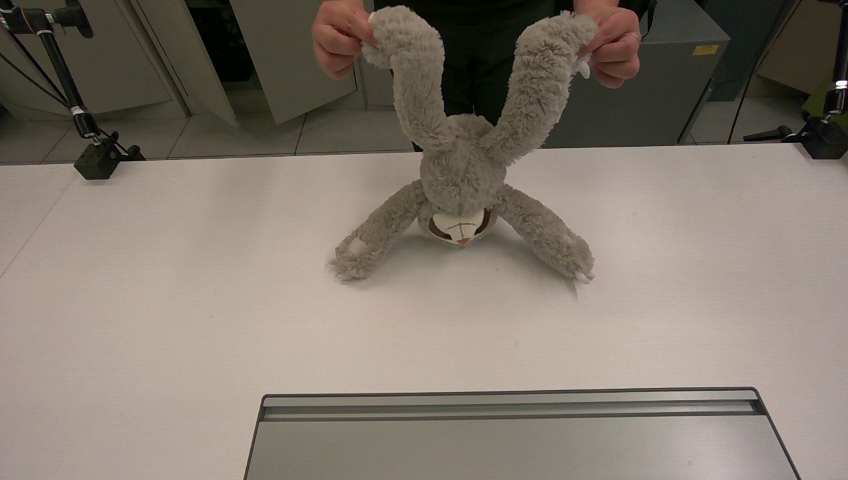} \\
    \qrow{\modelname (Ours)} & \qimg{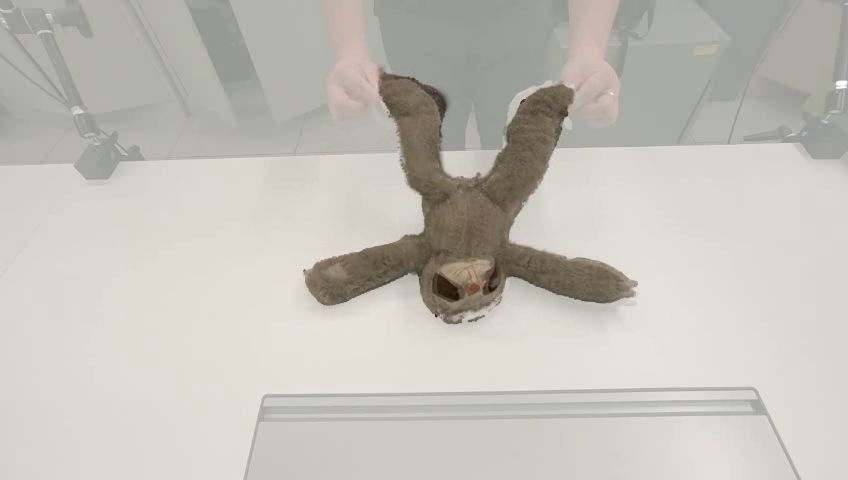}     & \qimg{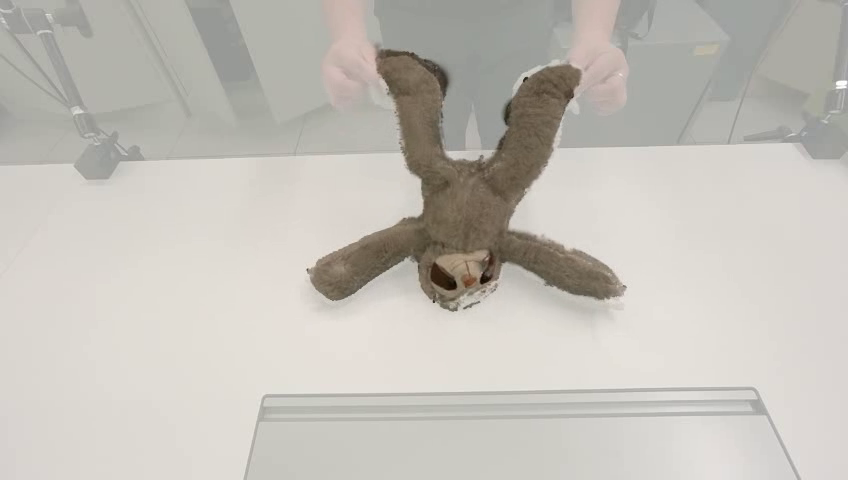}
                             & \qimg{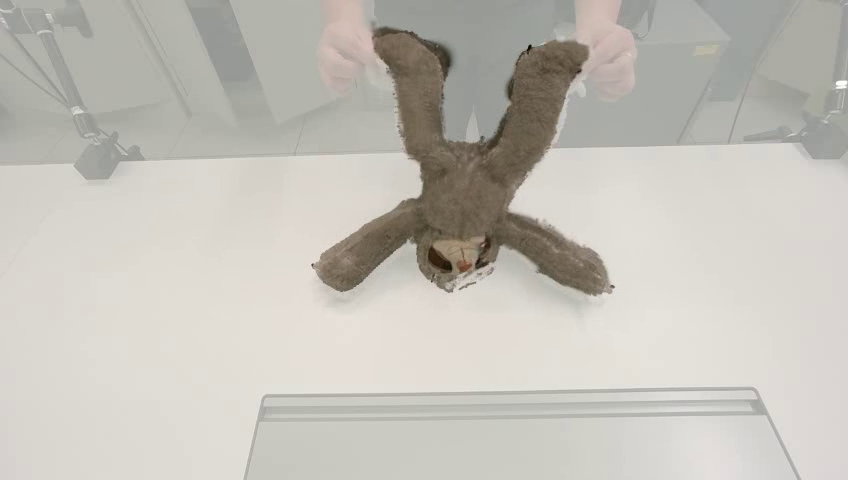}     & \qimg{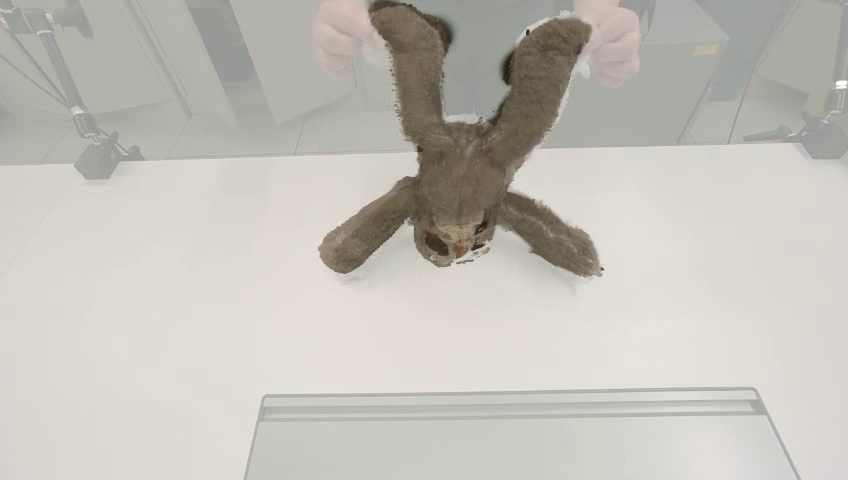} \\
    \qrow{PhysTwin}          & \qimg{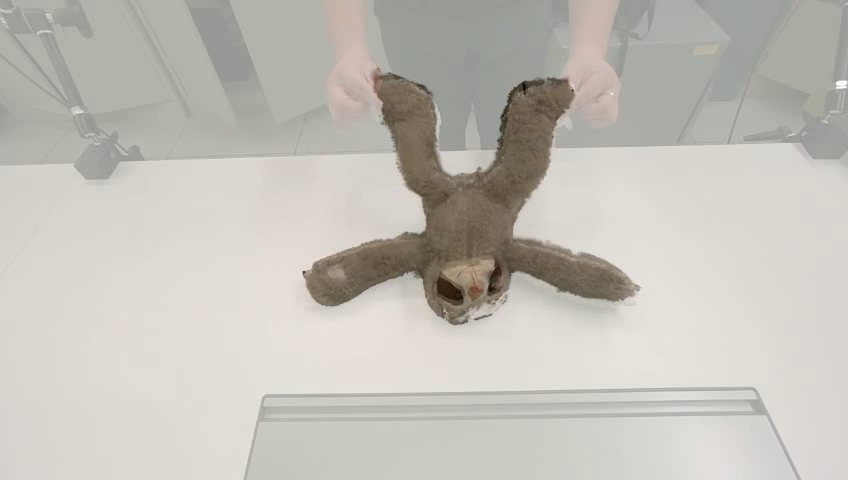} & \qimg{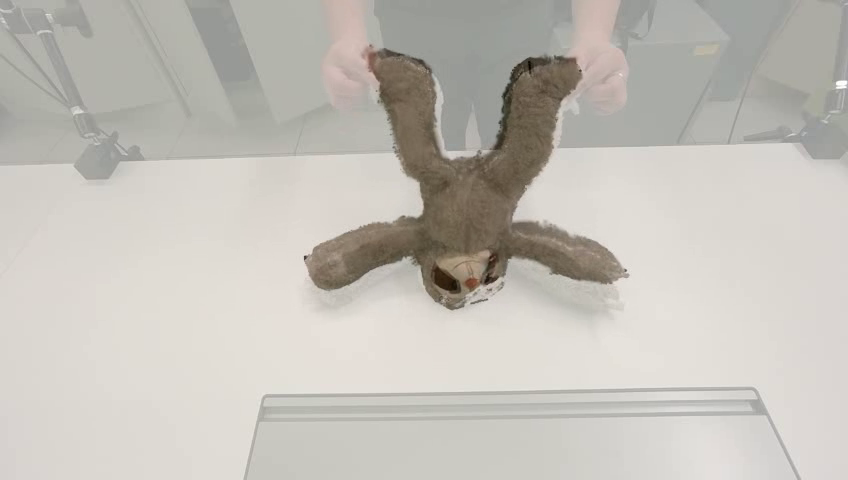}
                             & \qimg{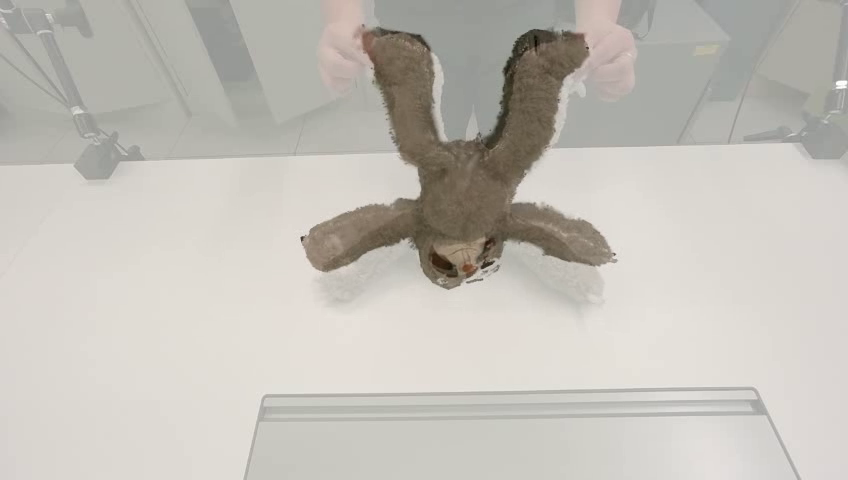} & \qimg{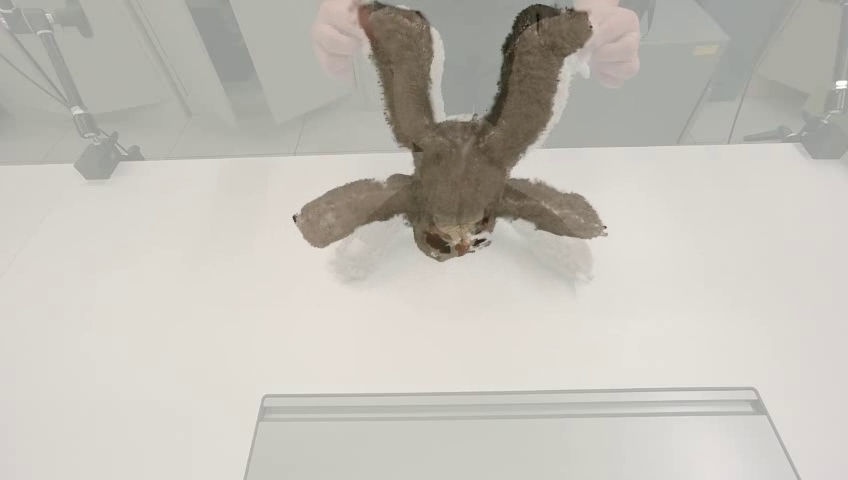} \\[4pt]

    \qrow{Observation}       & \qimg{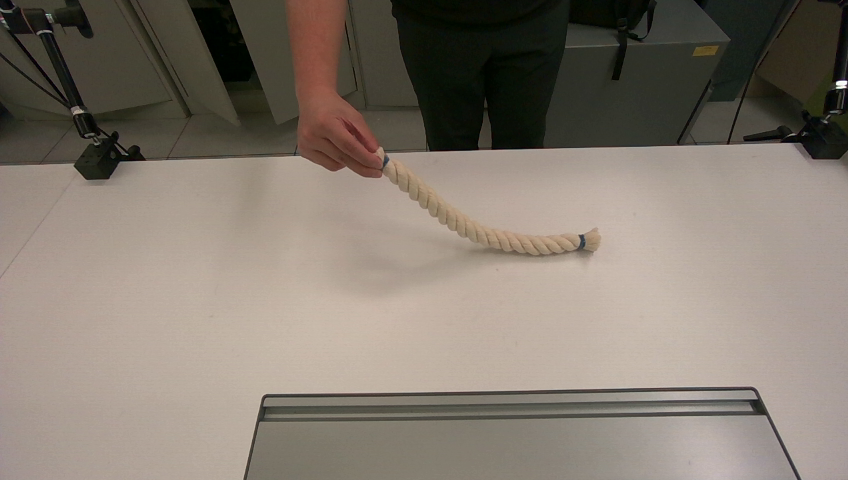}        & \qimg{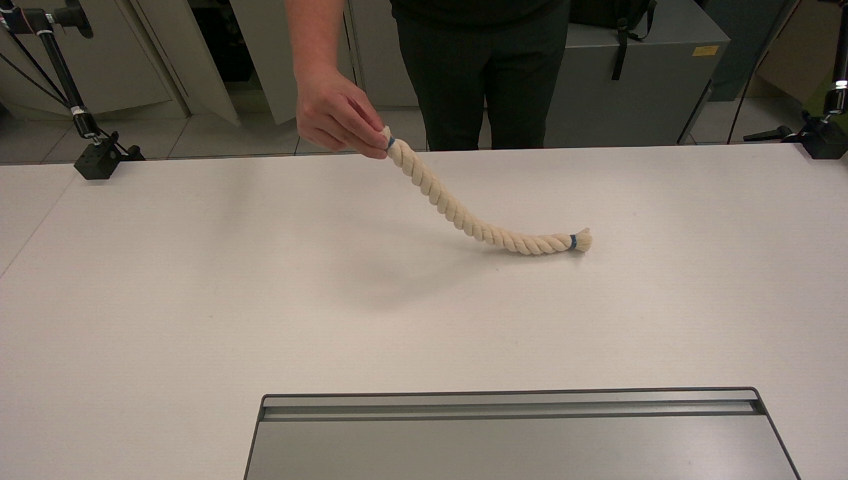}
                             & \qimg{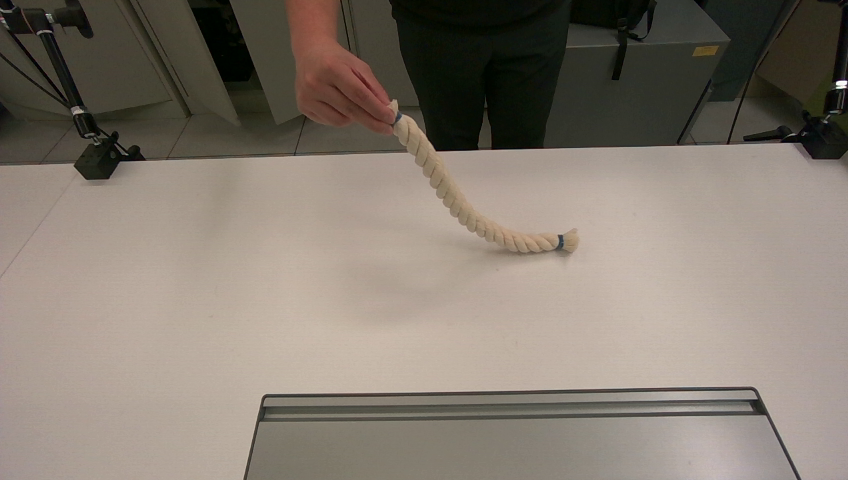}        & \qimg{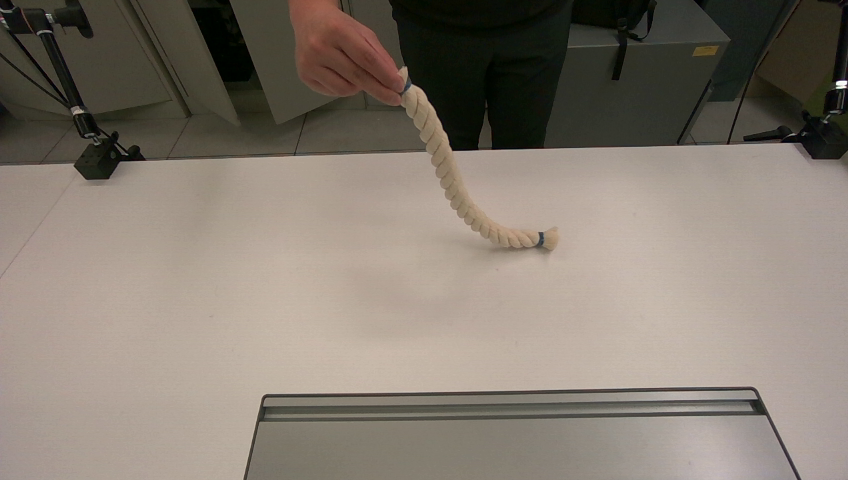} \\
    \qrow{\modelname (Ours)} & \qimg{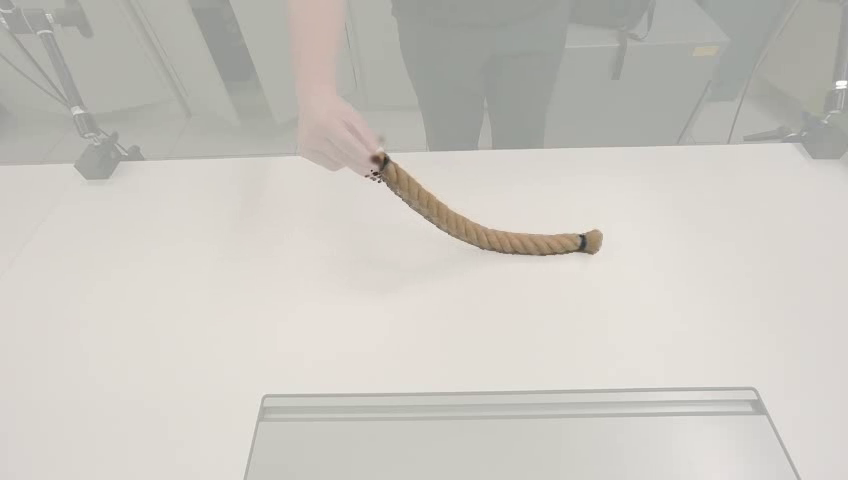}      & \qimg{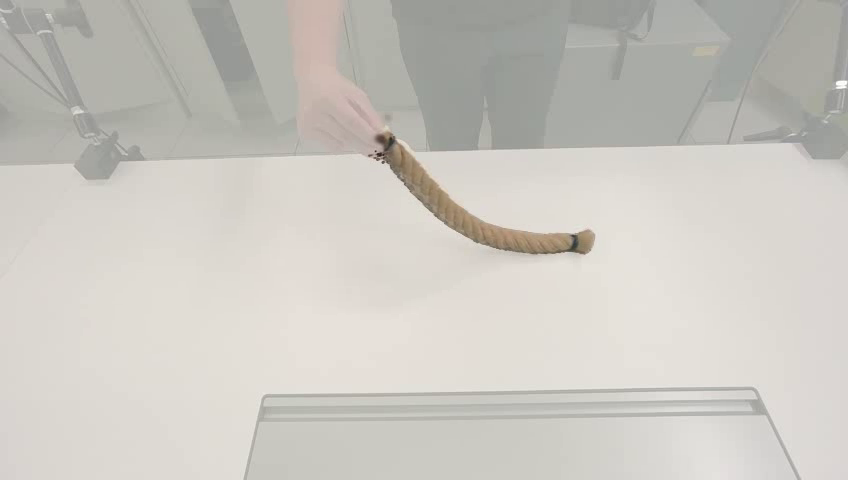}
                             & \qimg{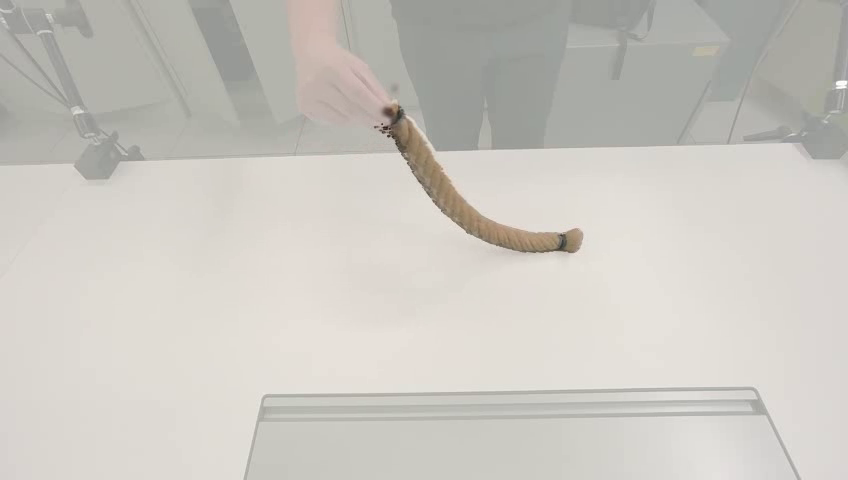}      & \qimg{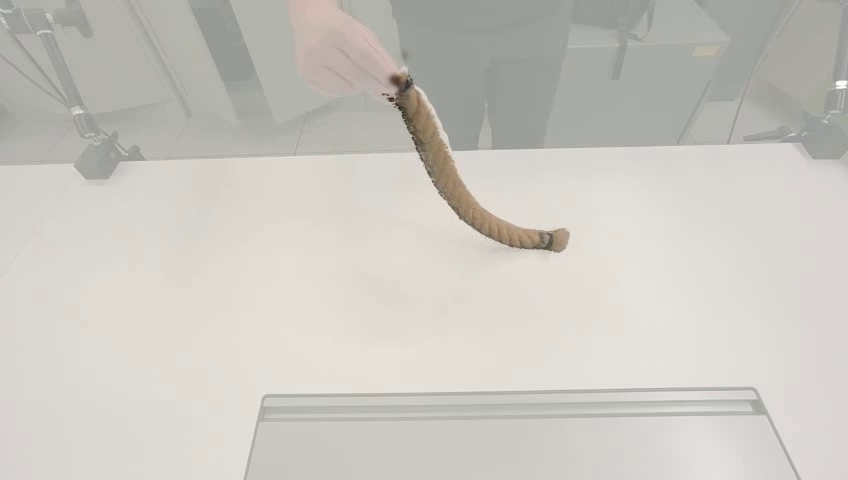} \\
    \qrow{PhysTwin}          & \qimg{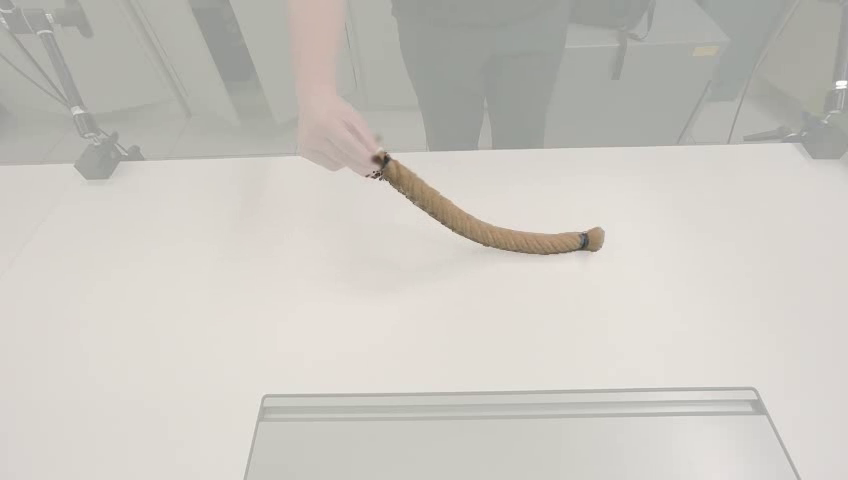}  & \qimg{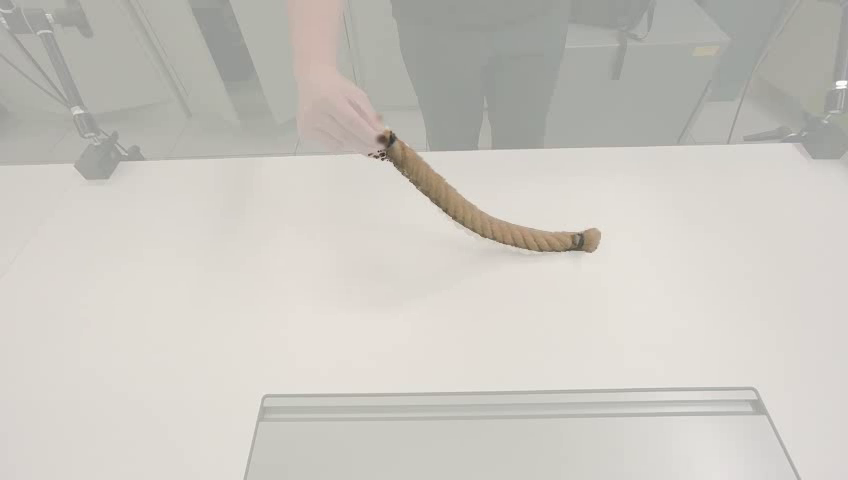}
                             & \qimg{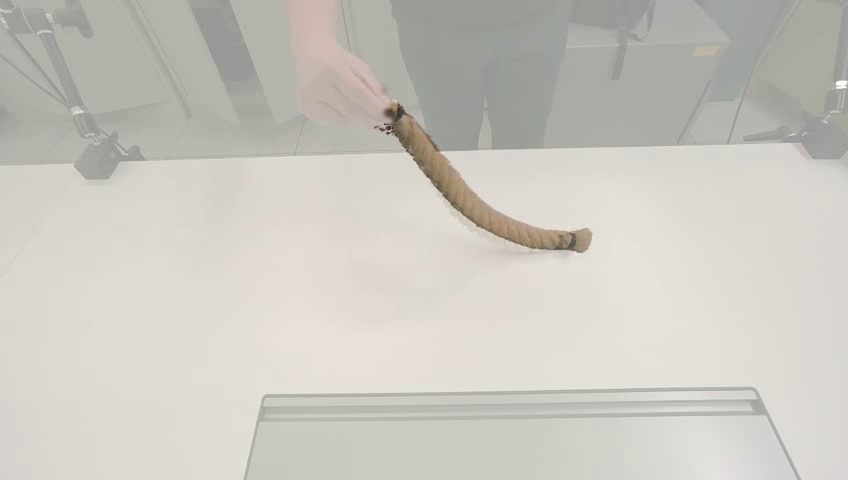}  & \qimg{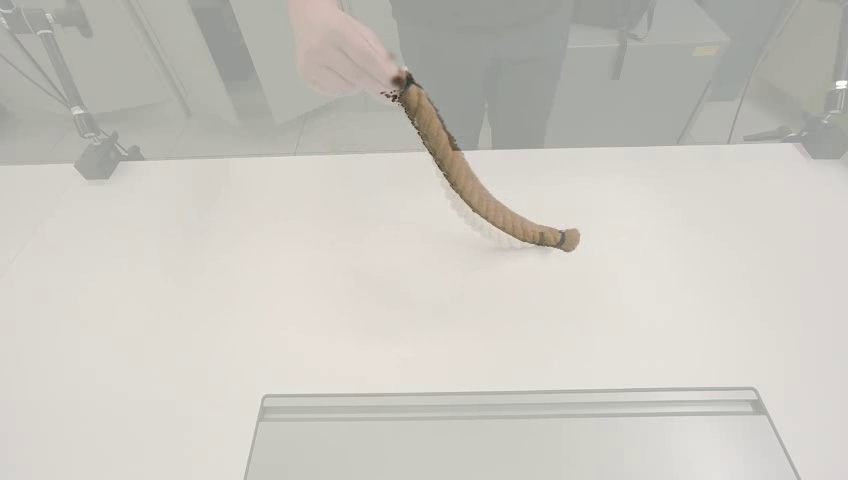} \\
  \end{tabular}
  \caption{\textbf{Qualitative results on Reconstruction \& Resimulation and Future Prediction.}
  For each scene we show four sampled frames: the left two come from the training window (reconstruction
  \& resimulation), and the right two are unseen future frames (future prediction). Rows compare the observation, our method (\textsc{\modelname}), and \textsc{PhysTwin}~\cite{jiang2025phystwin}. Our method tracks the deformation faithfully inside the training window and produces more stable and physically
  consistent rollouts beyond it.}
  \label{fig:qual_comparison}
\end{figure*}
 \subsection{Model Architecture}
 The goal of the model is to learn a transferable material-to-physics mapping for spring-mass simulation. The predictor uses two decoders at different physical scopes. An edge-level spring decoder estimates stiffness for each spring from material, motion, and local geometry cues, because stiffness controls local deformation and can vary across parts and edges. A global decoder estimates damping and contact
parameters from object-level material and motion cues, since these parameters govern overall energy dissipation and interaction response. This separation matches the structure of the simulator while allowing local material heterogeneity to affect spring behavior.

  Given the part decomposition $\mathcal{P}$ and spring graph $\mathcal{E}$, each part $P_m$ has an MLLM-queried material class prior $\mathbf{m} \in \mathbb{R}^{N_{\mathrm{mat}}}$ over $N_{\mathrm{mat}}$ predefined material classes, where $\mathbf{m}_q\ge 0$ and $\sum_{q=1}^{N_{\mathrm{mat}}}\mathbf{m}_q=1$. We introduce a learnable material codebook
  \begin{equation}
  \mathcal{C}=\{\mathbf{e}_q\}_{q=1}^{N_{\mathrm{mat}}},
  \qquad
  \mathbf{e}_q \in \mathbb{R}^{d_e},
  \end{equation}
  where each entry represents a latent material concept, and the part material embedding is defined as 
  \begin{equation}
  \mathbf{z}^{\mathrm{mat}}=\sum_{q=1}^{N_{\mathrm{mat}}}
  \mathbf{m}_q\mathbf{e}_q .
  \end{equation}
  For an edge $(i,j)$, its material feature $\mathbf{z}^{\mathrm{mat}}_{ij}$ is derived from the embeddings of the endpoint parts: intra-part edges use the embedding of their part, while boundary edges use the
  average embedding of the two adjacent parts.

  We also extract a video feature $\V{z}^{\mathrm{vid}}$ from the monocular video $\mathcal{V}=\{I_t\}_{t=1}^{T}$ using a pretrained video encoder along with a lightweight projector
  \begin{equation}
  \V{z}^{\mathrm{vid}} =f_{\mathrm{vid}}(\mathcal{V}),
  \end{equation}
  and compute an edge geometry feature
  \begin{equation}
  \mathbf{z}^{\mathrm{geo}}_{ij}=f_{\mathrm{geo}}(\mathbf{g}_{ij}),
  \end{equation}
  where $\mathbf{g}_{ij}$ encodes the rest length, relative displacement, edge direction, and local neighborhood statistics of spring $(i,j)$.

  The edge-level decoder predicts stiffness for each object spring:
  \begin{equation}
  \hat{k}_{ij}
  =
  f_{\mathrm{spring}}
  \left(
  \mathbf{z}^{\mathrm{vid}},
  \mathbf{z}^{\mathrm{mat}}_{ij},
  \mathbf{z}^{\mathrm{geo}}_{ij}
  \right).
  \end{equation}
  Thus, stiffness is edge-specific: material features provide shared part-level context, while local geometry allows variation within and across parts.

  The global decoder predicts object-level damping and contact parameters:
  \begin{equation}
  (\hat{\gamma}, \hat{\delta}, \hat{\mu}, \hat{\epsilon})
  =
  f_{\mathrm{global}}
  \left(
  \mathbf{z}^{\mathrm{vid}},
  \mathbf{z}^{\mathrm{mat}}_{\mathrm{global}}
  \right),
  \end{equation}
  where $\hat{\gamma}$ is spring damping, $\hat{\delta}$ is global drag damping, and $(\hat{\mu},\hat{\epsilon})$ denote contact friction and collision elasticity. The global material feature $\mathbf{z}
  ^{\mathrm{mat}}_{\mathrm{global}}$ is aggregated from part material embeddings. These parameters are shared at the object level because they describe global dissipation and contact response rather than individual spring connectivity.

  \subsection{Training and Inference}
\begin{figure*}[t!]
  \centering
  \setlength{\tabcolsep}{0.6pt}
  \renewcommand{\arraystretch}{0}
  \setlength{\qcellw}{0.205\textwidth}

  \newcommand{\qimg}[1]{\includegraphics[width=\qcellw]{#1}}
  \newcommand{\qrow}[1]{\rotatebox{90}{\scriptsize #1}}

  \begin{tabular}{c@{\hskip 1pt}cccc}

  & \multicolumn{4}{l}{\textcolor{red!70!black}{\small\itshape Future Prediction} 
  \hspace{0.5em}\rule[0.5ex]{0.68\textwidth}{0.4pt}\,$>$} \\[-2pt]
  & \multicolumn{4}{c}{\small\bfseries Unseen Interaction} \\[2pt]

    \qrow{Observation}       & \qimg{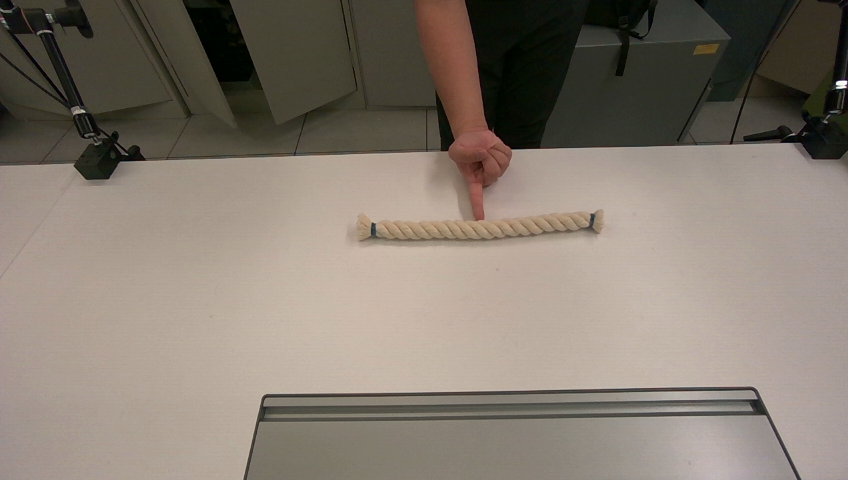}       & \qimg{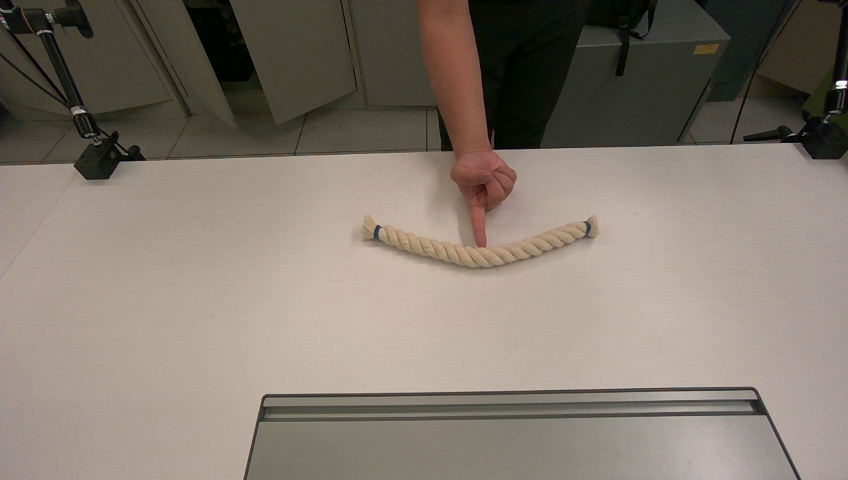}
                             & \qimg{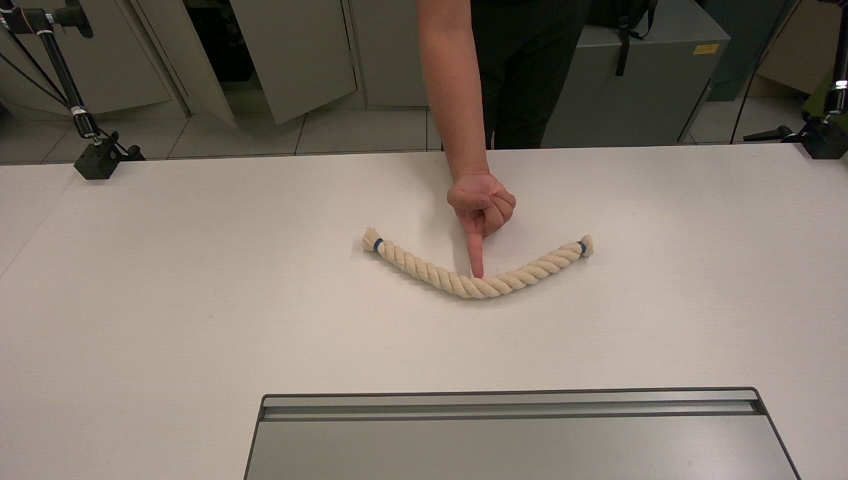}       & \qimg{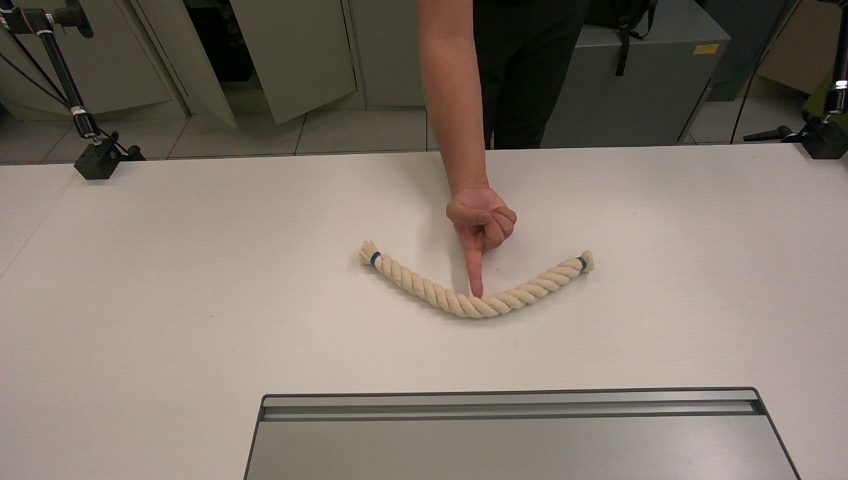} \\
    \qrow{\modelname (Ours)} & \qimg{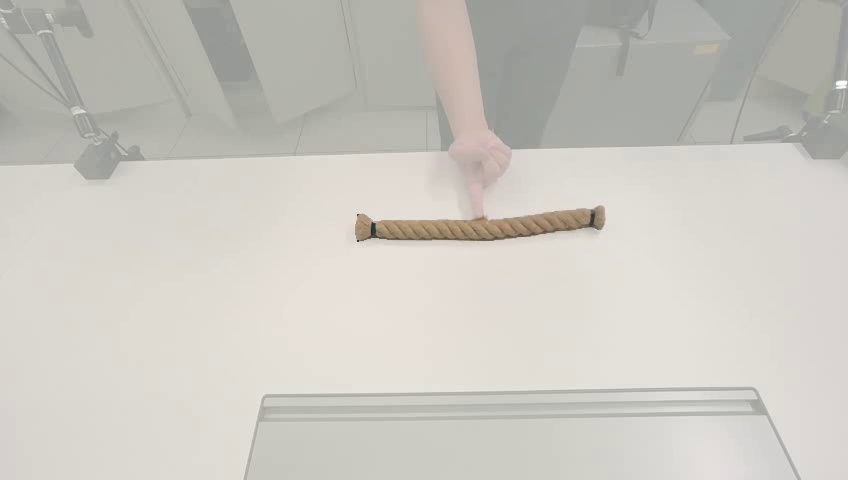}     & \qimg{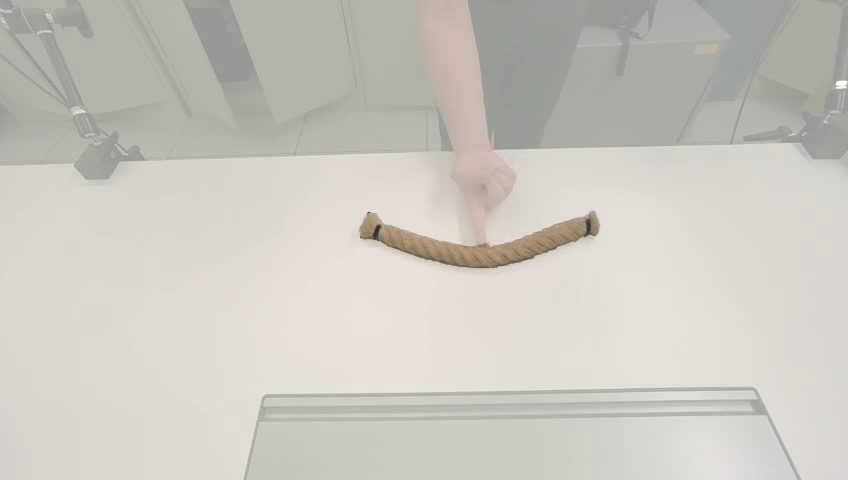}
                             & \qimg{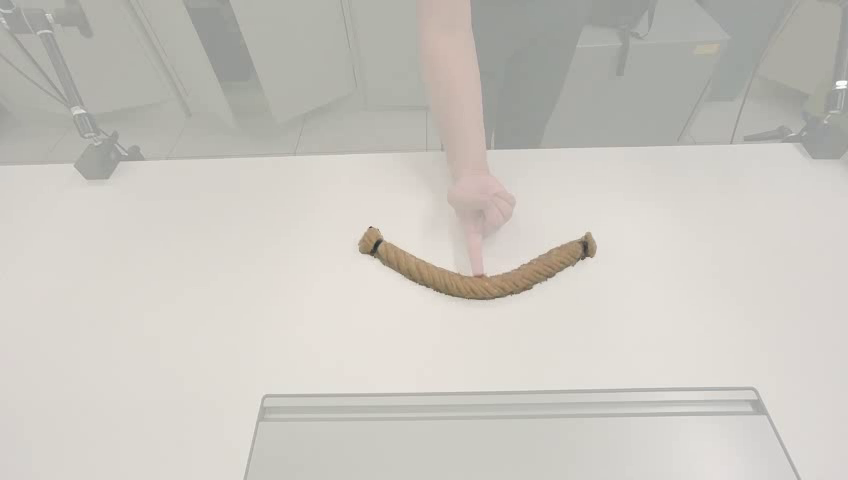}     & \qimg{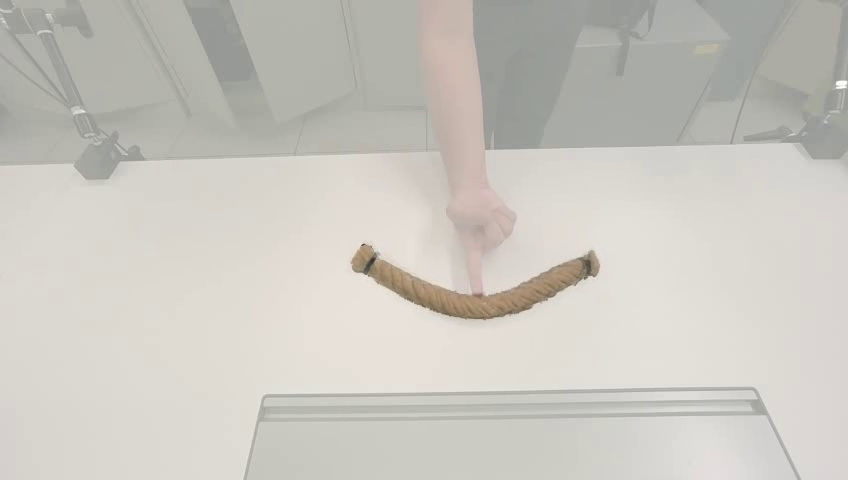} \\
    \qrow{PhysTwin}          & \qimg{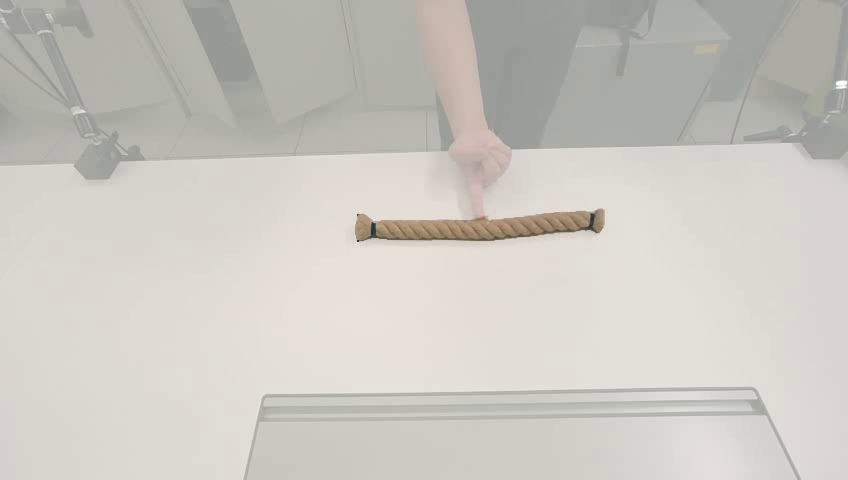} & \qimg{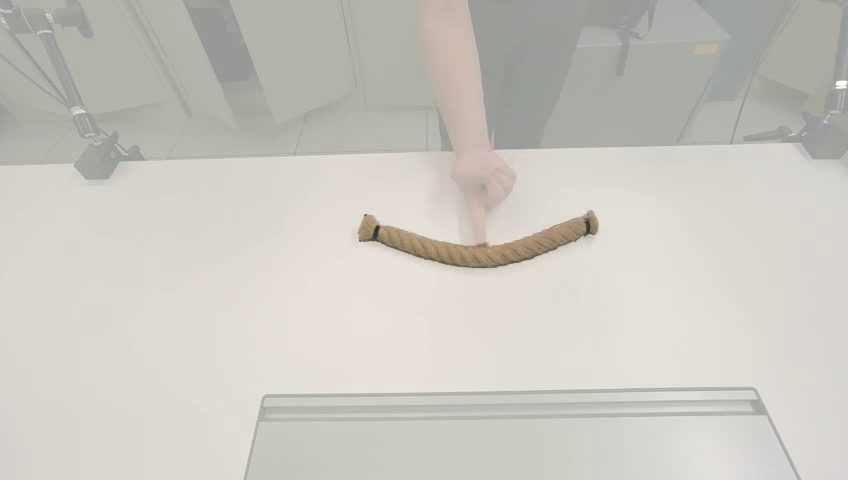}
                             & \qimg{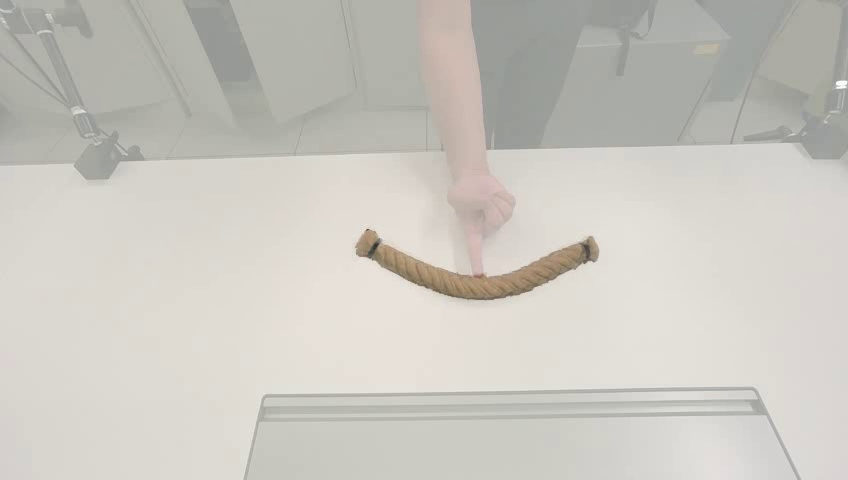} & \qimg{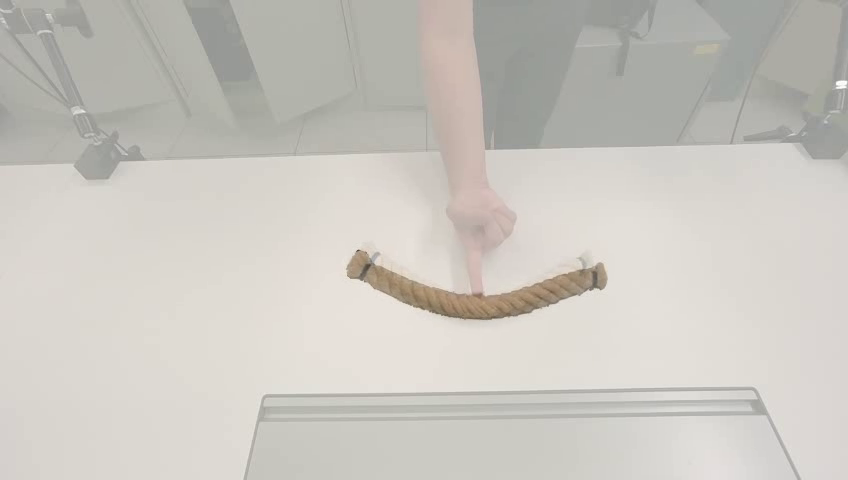} \\[2pt]

    \qrow{Observation}       & \qimg{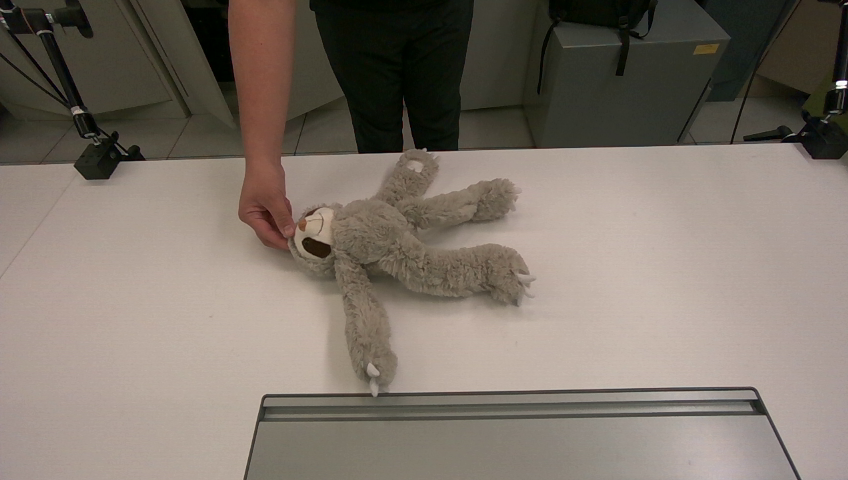}       & \qimg{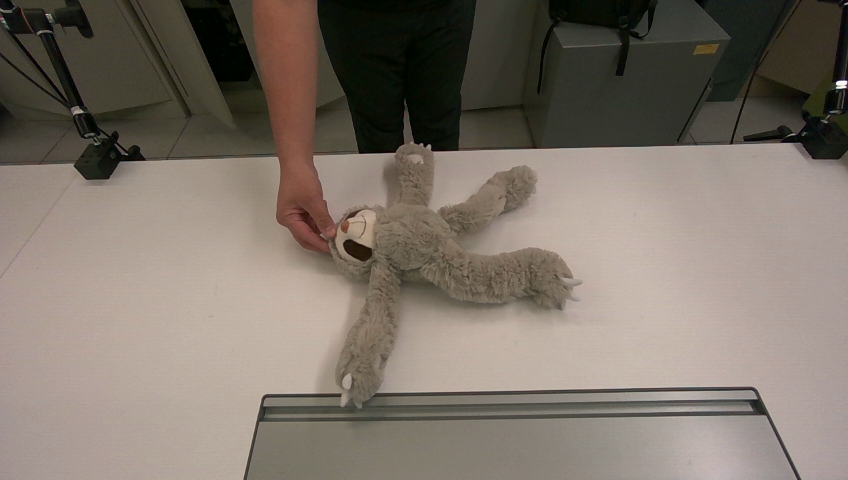}
                             & \qimg{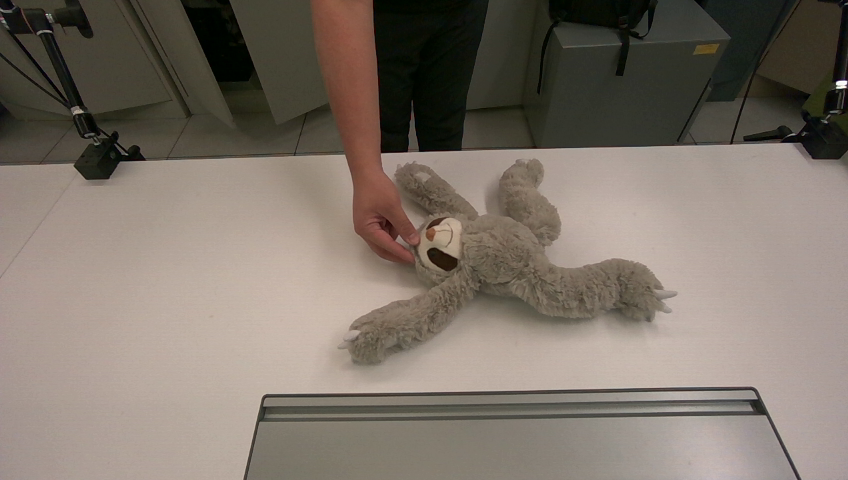}       & \qimg{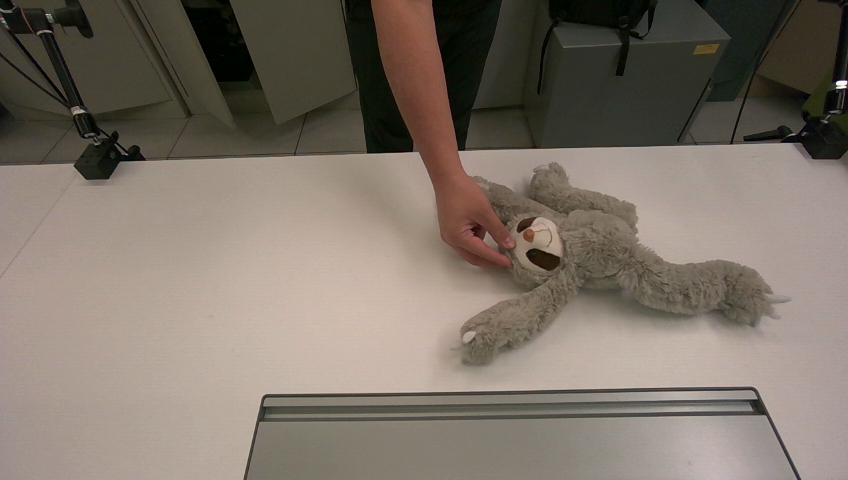} \\
    \qrow{\modelname (Ours)} & \qimg{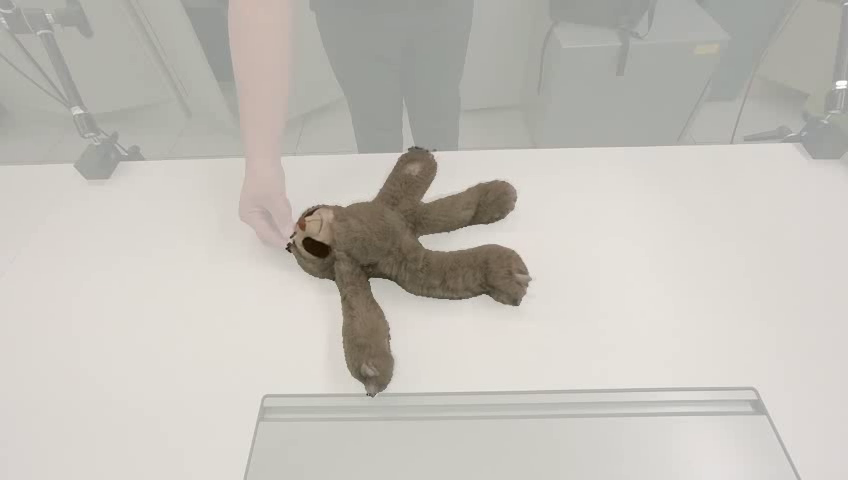}     & \qimg{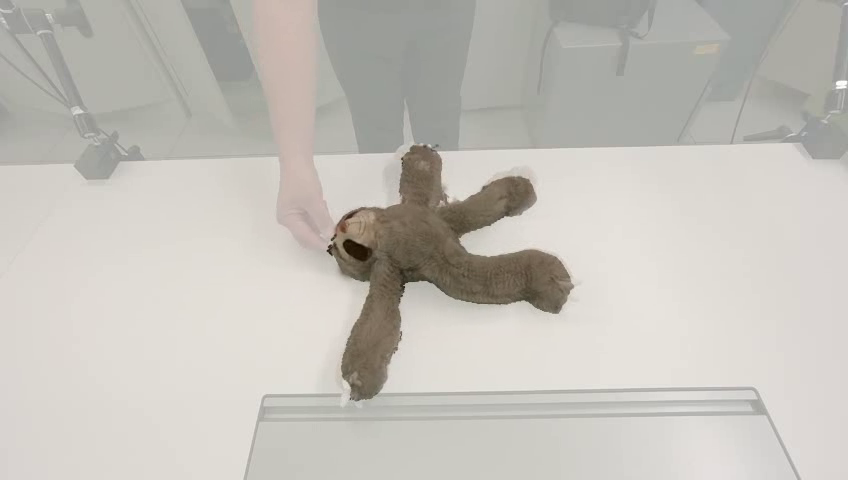}
                             & \qimg{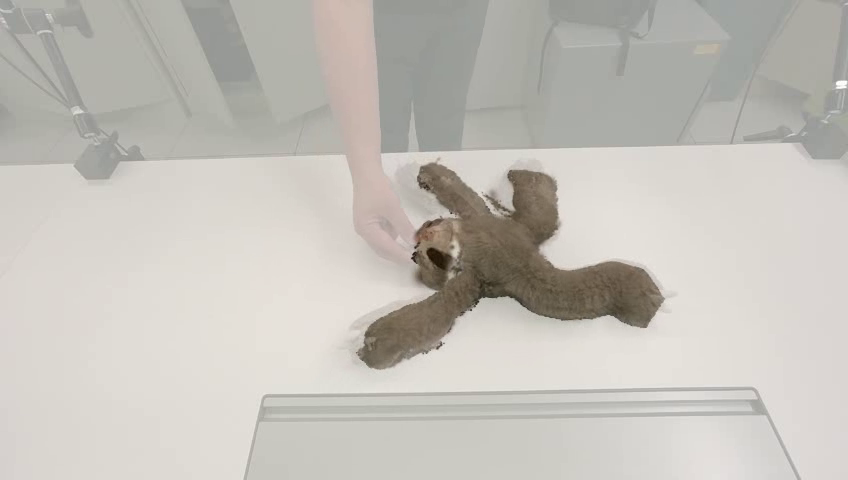}     & \qimg{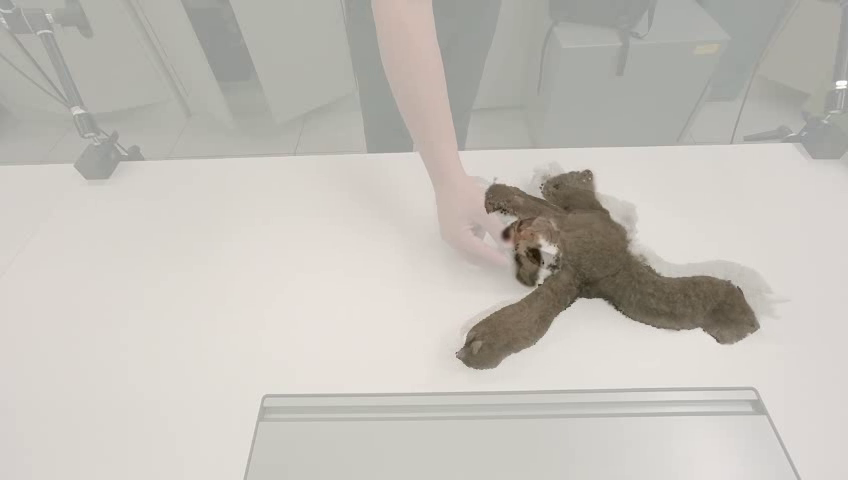} \\
    \qrow{PhysTwin}          & \qimg{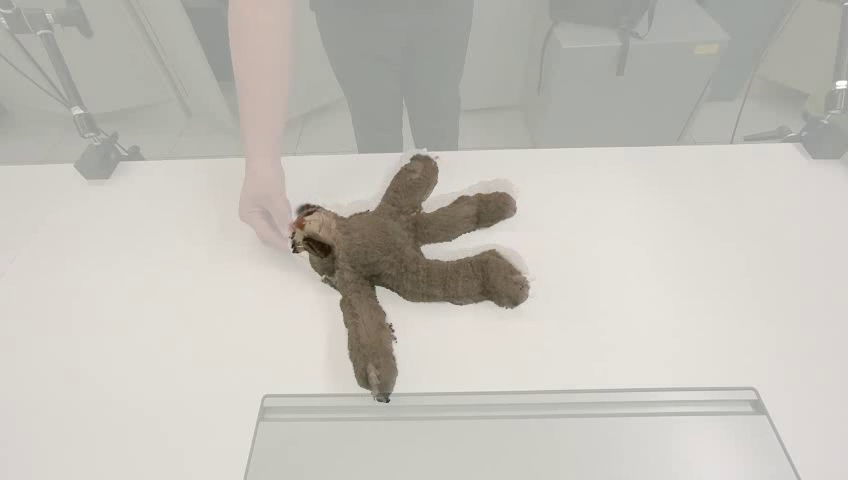} & \qimg{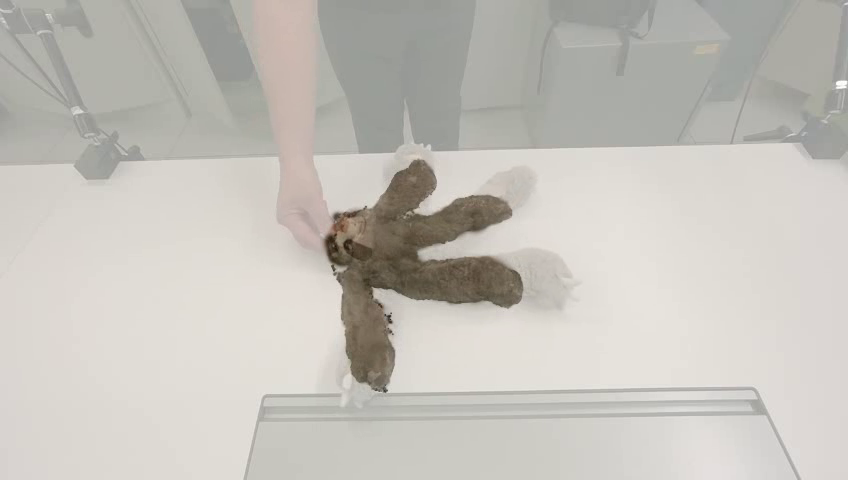}
                             & \qimg{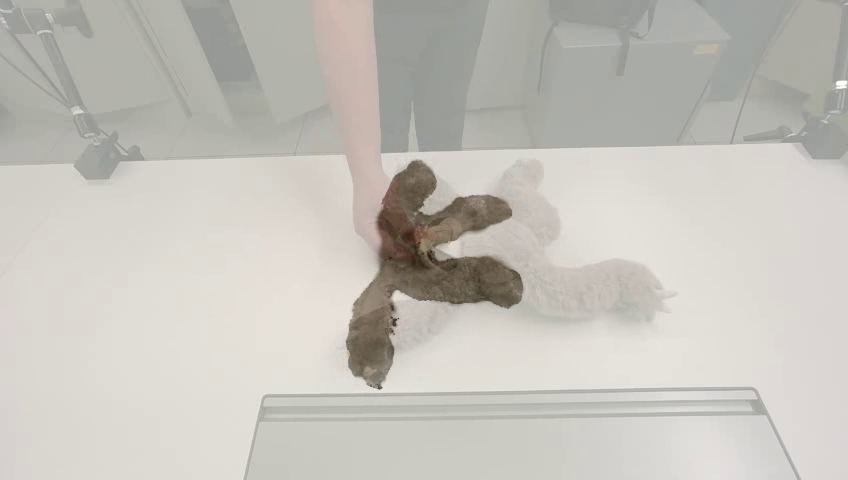} & \qimg{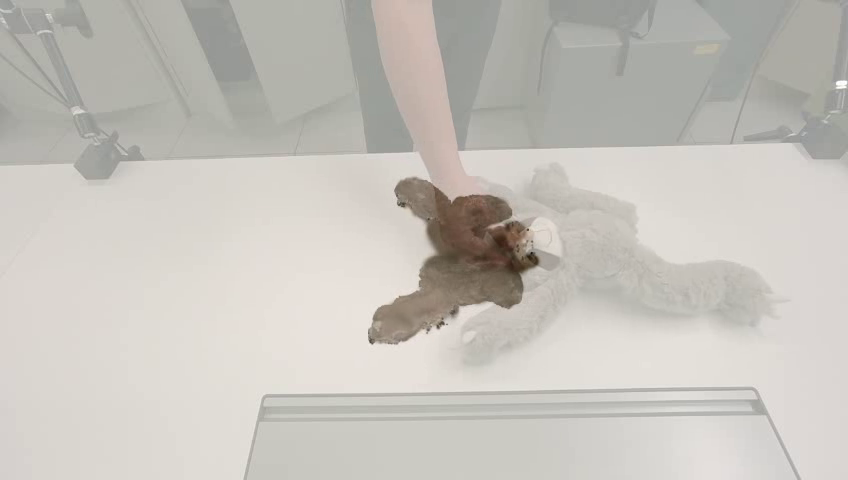} \\[3pt]

    \multicolumn{5}{c}{\small\textbf{Unseen Object}} \\[2pt]

    \qrow{Observation}       & \qimg{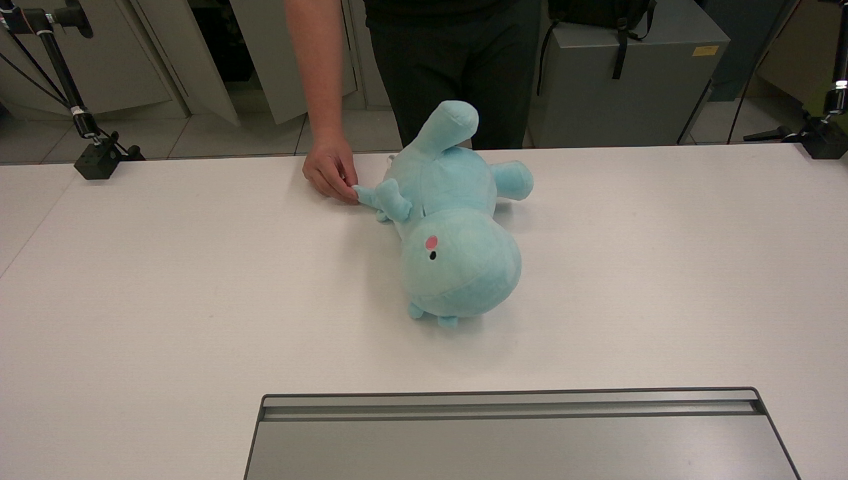}       & \qimg{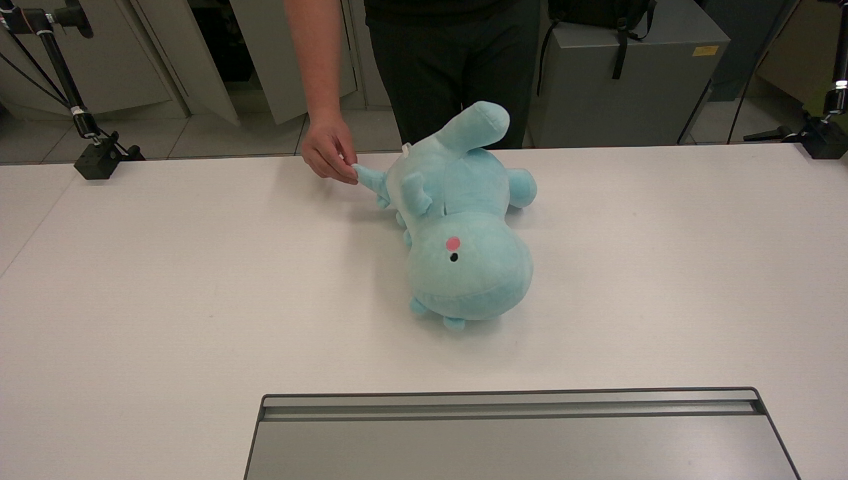}
                             & \qimg{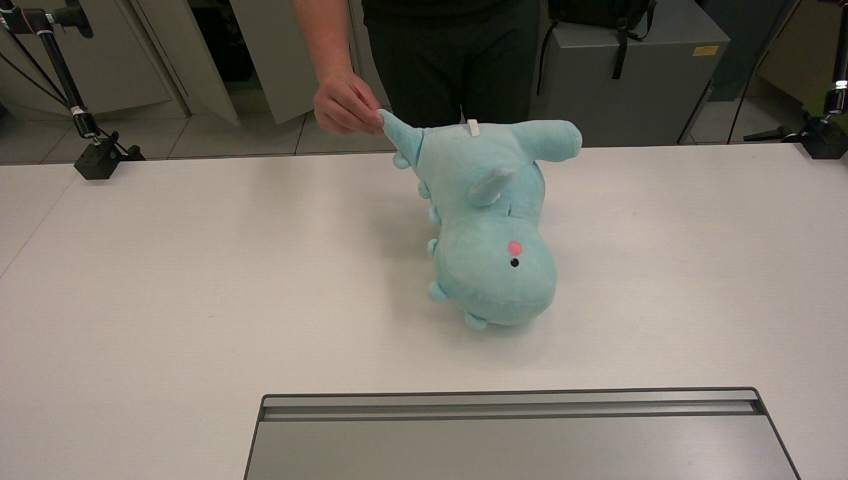}       & \qimg{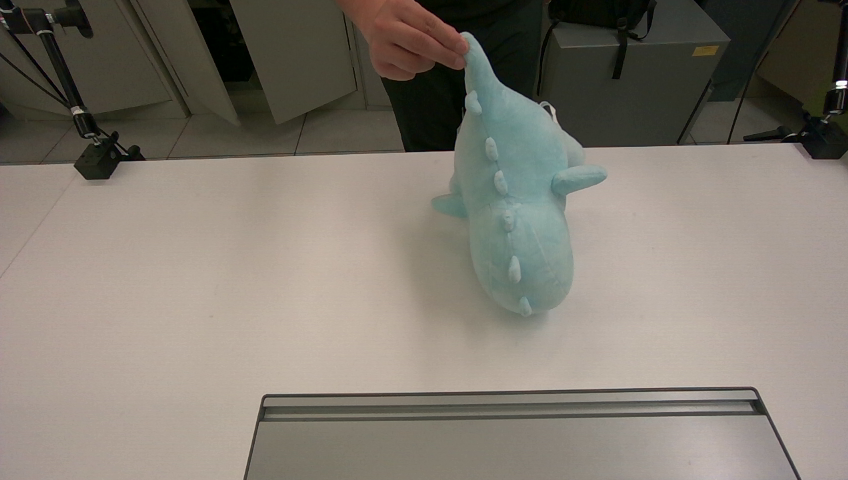} \\
    \qrow{\modelname (Ours)} & \qimg{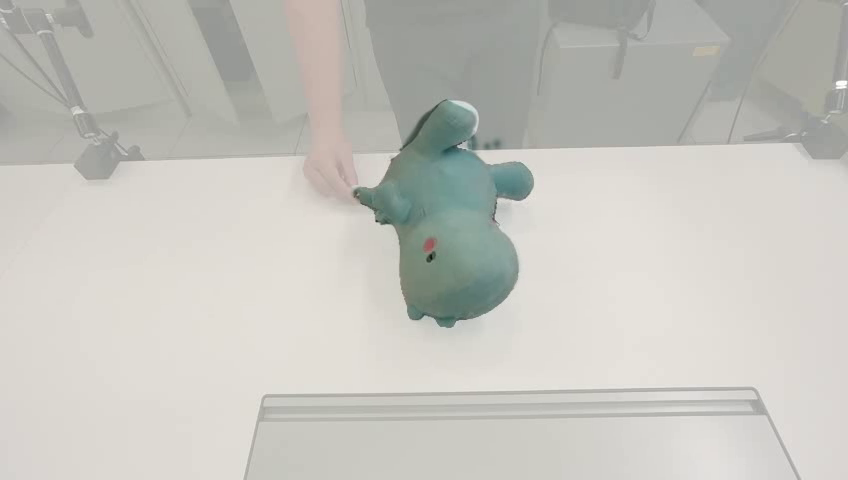}     & \qimg{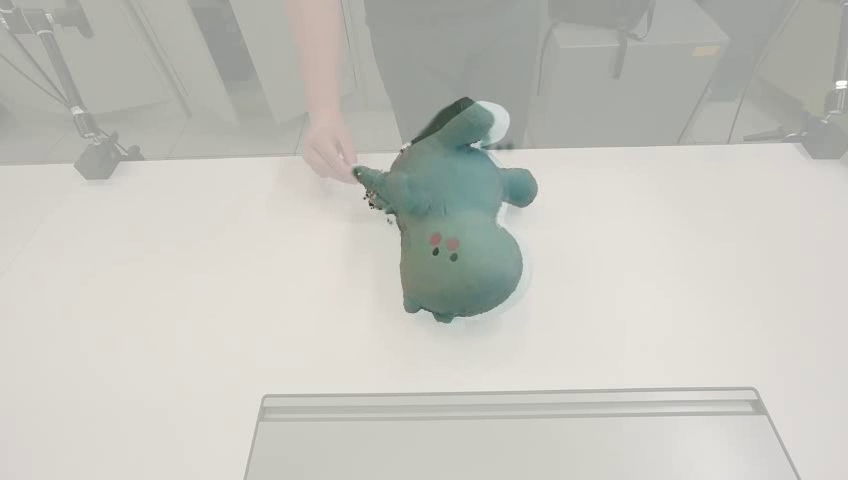}
                             & \qimg{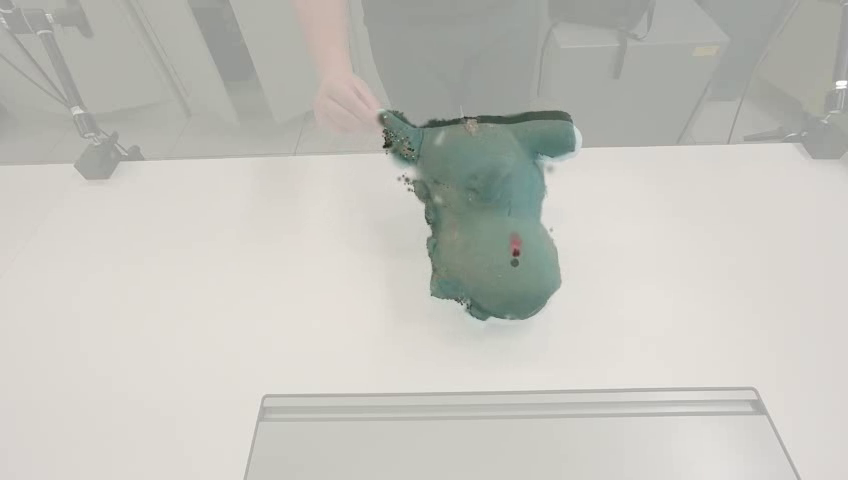}     & \qimg{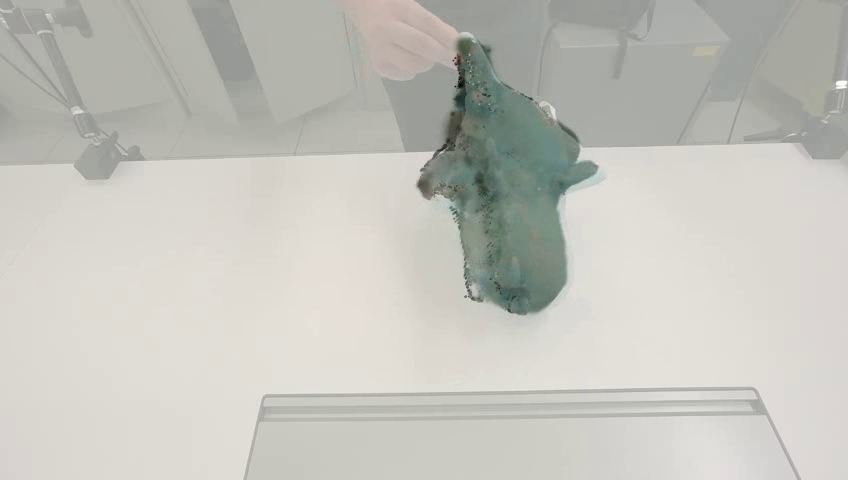} \\

  \end{tabular}

  \vspace{-0.5em}
  \caption{\textbf{Qualitative results on future prediction under unseen interaction and unseen object.}
  We show four sampled future frames for each example.
  The top two examples evaluate generalization to \emph{unseen interaction}, where the object is subjected to interaction patterns not observed during optimization.
  The bottom example evaluates generalization to \emph{unseen object}, where the model is applied to a novel object.
  Rows compare the observation, our method (\textsc{\modelname}), and \textsc{PhysTwin}~\cite{jiang2025phystwin}.
  Our method produces more accurate deformation trajectories.}
  \label{fig:qual_unseen}
\end{figure*}
  During training, the predicted parameters instantiate the spring-mass simulator together with the part-aware topology $\mathcal{E}$. Starting from the initial Gaussian centers, the simulator rolls out the
  mass-point trajectories and produces the deformed Gaussian object over time. We optimize the predictor by matching the simulated rollout to the observed object motion and geometry:
  \begin{equation}
  \mathcal{L}_{\mathrm{sim}}
  =
  \lambda_{\mathrm{trk}}\mathcal{L}_{\mathrm{trk}}
  +
  \lambda_{\mathrm{cham}}\mathcal{L}_{\mathrm{cham}} .
  \end{equation}
  Here, $\mathcal{L}_{\mathrm{trk}}$ encourages simulated mass-point trajectories to follow observed or pseudo 3D motion, while $\mathcal{L}_{\mathrm{cham}}$ matches the simulated object geometry to the
  observed geometry over time.

  The queried physical prior is used as a part-level reference distribution rather than direct supervision for every edge. For each part, we aggregate the predicted spring statistics and regularize them toward the physical range implied by the MLLM material prior:
  \begin{equation}
  \mathcal{L}_{\mathrm{prior}}
  =
  \sum_m
  D\left(
  \hat{p}_m(\Theta)
  \,\|\,
  p_m^{\mathrm{prior}}(\Theta)
  \right),
  \end{equation}
  where $\hat{p}_m(\Theta)$ denotes the predicted part-level parameter statistics and $p_m^{\mathrm{prior}}(\Theta)$ denotes the queried reference distribution. The full objective is
  \begin{equation}
  \mathcal{L}
  =
  \mathcal{L}_{\mathrm{sim}}
  +
  \lambda_{\mathrm{prior}}\mathcal{L}_{\mathrm{prior}} .
  \end{equation}

  At inference time, we reconstruct the Gaussians, decompose them into semantic parts,  construct the part-aware topology, and predict all simulator parameters with a single forward pass from a monocular video.
\begin{table*}[t]
\centering
\caption{\textbf{Quantitative comparison on reconstruction/resimulation and future prediction.} The left part evaluates performance within the observed interaction window, while the right part measures temporal extrapolation to unseen future frames.}
\label{tab:main_quantitative}
\resizebox{\textwidth}{!}{
\begin{tabular}{l cccccc cccccc}
\toprule
Task
& \multicolumn{6}{c}{Reconstruction \& Resimulation}
& \multicolumn{6}{c}{Future Prediction} \\
\cmidrule(lr){2-7} \cmidrule(lr){8-13}
Method & CD$\downarrow$ & Track Error$\downarrow$ & IoU \%$\uparrow$ & PSNR$\uparrow$
& SSIM$\uparrow$ & LPIPS$\downarrow$ & CD$\downarrow$ & Track Error$\downarrow$ & IoU\%$\uparrow$ & PSNR$\uparrow$ & SSIM$\uparrow$ & LPIPS$\downarrow$ \\
\midrule
Spring-Gaus & 0.041 & 0.050 & 57.6 & 23.445 & 0.928 & 0.102 & 0.062 & 0.094 & 46.4 & 22.488 & 0.924 & 0.113 \\
GS-Dynamics & 0.014 & 0.022 & 72.1 & 26.260 & 0.940 & 0.052 & 0.041 & 0.070 & 49.8 & 22.540 & 0.924 & 0.097 \\
PhysTwin    & 0.005 & 0.009 & 84.4& 28.214 & 0.945 & 0.034 & 0.012 & 0.022 & 72.5& 25.617 & 0.941 & 0.055 \\
NeuSpring  & \textbf{0.004} & \textbf{0.008} & \textbf{86.2}& \textbf{28.664} & \textbf{0.947} & \textbf{0.031} & 0.009 & 0.017 & 75.9 & 26.361 & 0.944 & 0.049 \\
Ours & 0.006 & 0.009 &84.1 & 28.376 &0.943 &0.036 & \textbf{0.008} & \textbf{0.015} & \textbf{79.8} & \textbf{26.724} & \textbf{0.946} & \textbf{0.046} \\
\bottomrule
\end{tabular}
}
\end{table*}

\begin{table*}[t]
\centering
\caption{
\textbf{Ablation study.} 
\textit{w/o part decomposition} removes part-level material modeling and predicts object-level physical parameters.
\textit{w/o $\mathcal{L}_{\mathrm{prior}}$} removes the VLM-guided material prior regularization.
\textit{w/o codebook} directly regresses physical parameters without the learned material codebook.
}
\label{tab:ablation}
\resizebox{\textwidth}{!}{
\begin{tabular}{l cccccc cccccc}
\toprule
Task
& \multicolumn{6}{c}{Reconstruction \& Resimulation}
& \multicolumn{6}{c}{Future Prediction} \\
\cmidrule(lr){2-7} \cmidrule(lr){8-13}
Method & CD$\downarrow$ & Track Error$\downarrow$ & IoU \%$\uparrow$ & PSNR$\uparrow$
& SSIM$\uparrow$ & LPIPS$\downarrow$ & CD$\downarrow$ & Track Error$\downarrow$ & IoU\%$\uparrow$ & PSNR$\uparrow$ & SSIM$\uparrow$ & LPIPS$\downarrow$ \\
\midrule
w/o codebook
   & 0.009 & 0.014 & 78.3 & 26.842 & 0.924 & 0.054
   & 0.012 & 0.023 & 73.5 & 25.106 & 0.925 & 0.069 \\
w/o $\mathcal{L}_{\mathrm{prior}}$
   & 0.007 & 0.011 & 82.0 & 27.812 & 0.937 & 0.042
   & 0.009 & 0.018 & 77.6 & 26.103 & 0.939 & 0.054 \\
w/o part decomposition
   & 0.008 & 0.012 & 80.7 & 27.431 & 0.932 & 0.047
   & 0.010 & 0.020 & 76.1 & 25.742 & 0.934 & 0.059 \\
Ours & \textbf{0.006} & \textbf{0.009} & \textbf{84.1} & \textbf{28.376} & \textbf{0.943} & \textbf{0.036}
   & \textbf{0.008} & \textbf{0.015} & \textbf{79.8} & \textbf{26.724} & \textbf{0.946} & \textbf{0.046} \\
\bottomrule
\end{tabular}
}
\end{table*}
\begin{table}[t]
\centering
\caption{\textbf{Quantitative comparison on generalization to unseen interaction.} We evaluate whether the estimated physical parameters can transfer to novel interaction patterns that are not observed during optimization.}
\label{tab:zero-shot}
\resizebox{\columnwidth}{!}{
\begin{tabular}{lcccccc}
\toprule
Method & CD$\downarrow$ & Track Error$\downarrow$ & IoU (\%)$\uparrow$ & PSNR$\uparrow$ & SSIM$\uparrow$ & LPIPS$\downarrow$ \\
\midrule
PhysTwin & 0.019 & 0.032 & 47.7 & 21.16 & 0.926 & 0.065 \\
NeuSpring & 0.014 & 0.029 & 50.2 & 21.78 & 0.928 & 0.063 \\
Ours  & \textbf{0.009}  & \textbf{0.014}  & \textbf{77.5} & \textbf{22.32} &\textbf{0.931}  & \textbf{0.059}  \\
\bottomrule
\end{tabular}
}
\end{table}

\section{Experiments}
\subsection{Experimental settings}
\paragraph{Datasets.}  We conduct training and quantitative evaluation on the PhysTwin dataset~\cite{jiang2025phystwin}, which contains 22 deformable-object interaction scenarios. Each scenario provides RGB-D videos of human interactions, where the annotated hand points are used as controller points to guide training. Following the standard split used in prior spring-mass baselines, we split each scenario's videos into training and test sets at a 7:3 ratio.

\paragraph{Baselines.}  We compare our method against four representative baselines: GS-Dynamics~\cite{gsdynamics}, Spring-Gaus~\cite{zhong2024springgaus}, PhysTwin~\cite{jiang2025phystwin}, and NeuSpring~\cite{xu2026neuspring}. Among them, GS-Dynamics is a learning-based neural dynamics method, while the others are physics-driven spring-mass approaches. In particular, PhysTwin and NeuSpring are the most closely related baselines to our setting, since both aim to recover simulation-ready deformable object models for reconstruction, resimulation, and future prediction.
\begin{figure*}[t!]
  \centering
  \includegraphics[width=\linewidth]{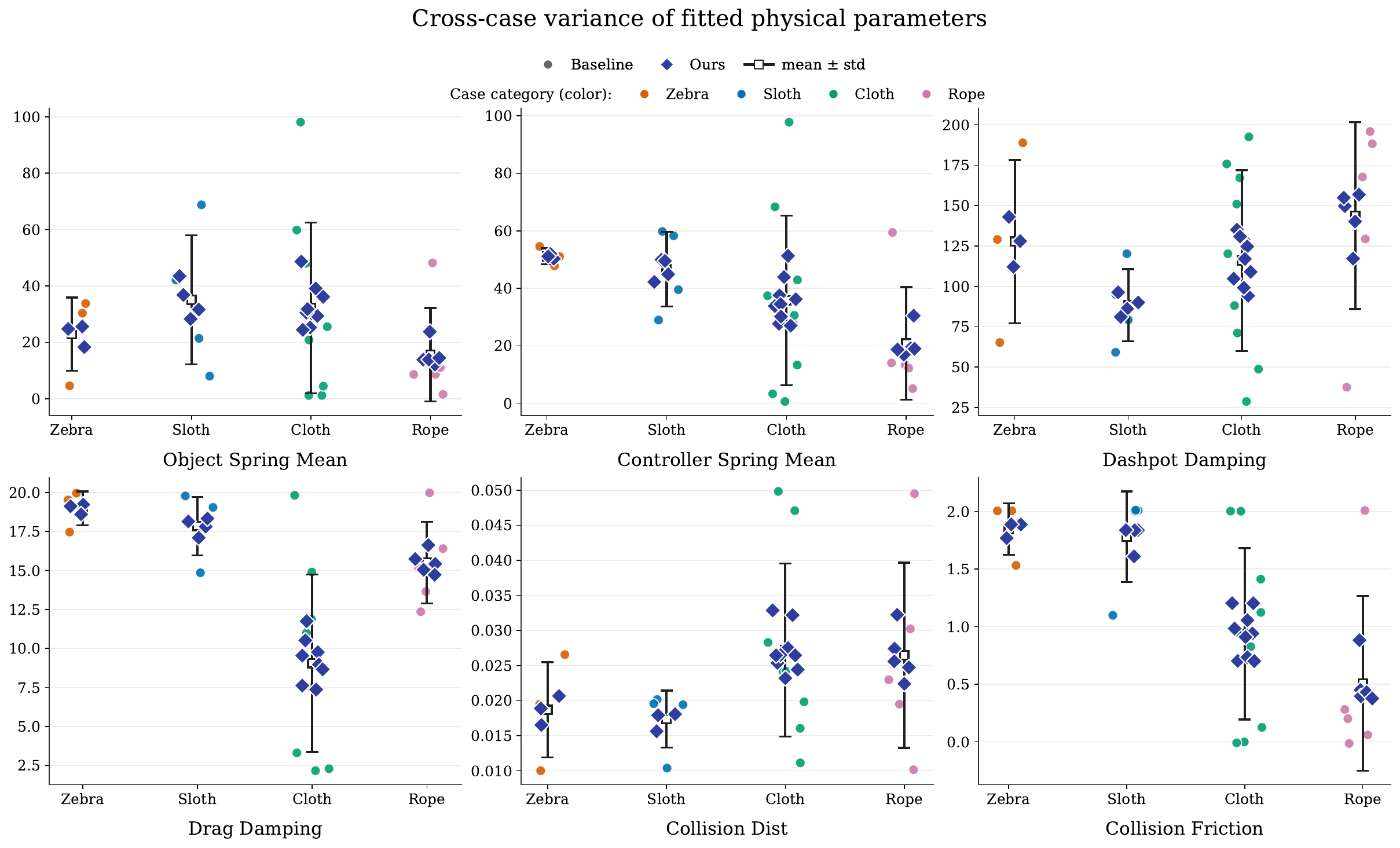}
  \caption{\textbf{Cross-case variance of fitted physical parameters for the same object category.} For each category (color-coded), circles denote per-case baseline estimates and squares with error bars denote the category mean ± standard deviation.  }
  \label{fig:cross-scene statics}
\end{figure*}
\paragraph{Metrics.}
We evaluate the predicted dynamics using both 3D and 2D metrics following prior spring-mass baselines. For 3D evaluation, we use Chamfer Distance (CD) and tracking error. For 2D evaluation, we use PSNR, SSIM, LPIPS, and silhouette IoU to measure the alignment between rendered predictions and observed image sequences. Following prior work, 2D evaluation is performed at the center viewpoint, and all metrics are averaged over frames and scenarios.
\paragraph{Implementation details.}
We use a frozen VideoMAE-base~\cite{tong2022videomae} encoder to extract motion features from $T=32$ monocular frames.
We use $K=5$ part segments and a 10-class material prior with a learnable codebook to provide material features.
Geometry, motion, and material features are fused by lightweight MLP decoders to predict spring-mass parameters.
We set $\lambda_{\mathrm{trk}}=1.0$, $\lambda_{\mathrm{cham}}=1.0$, and $\lambda_{\mathrm{prior}}=10^{-3}$.
\subsection{Results}
We evaluate our method under three settings: reconstruction \& resimulation, future prediction, and generalization to unseen interactions and objects.
Overall, our method achieves competitive performance on reconstruction and resimulation, while showing clear advantages in future prediction and generalization results.
\paragraph{Reconstruction \& resimulation}The left part of Table~\ref{tab:main_quantitative} reports the reconstruction and resimulation results on seen objects and scenes. We observe that our method achieves results comparable to per-case optimization baselines across most metrics. Although our method does not surpass all baselines on every metric, it remains consistently competitive, indicating that the proposed framework can recover physically plausible dynamics without overfitting the parameters to each case separately. Figure~\ref{fig:qual_comparison} shows consistent qualitative observations.
Compared with PhysTwin, our method produces more faithful deformation trajectories and more stable visual results under interactive manipulation.
\paragraph{Future prediction.}
Table~\ref{tab:main_quantitative} also reports the results on future frame prediction.
Compared with reconstruction and resimulation, future prediction is more challenging because the model must extrapolate the learned dynamics beyond the observed input window.
In this setting, our method achieves the best performance across all metrics.
This indicates that the learned physical parameters are not only able to explain the observed deformation, but also support more accurate temporal rollout to unseen future states.
The future prediction results in Figure~\ref{fig:qual_comparison} further validate this advantage.
\paragraph{Generalization to unseen objects and scenes.}
We also evaluate zero-shot transfer to unseen objects and scenes against baselines.
Existing baselines are mainly designed to fit each test case individually, and therefore do not naturally support zero-shot evaluation.
To make the comparison as fair as possible, for each baseline we estimate shared parameters by averaging the optimized parameters from $n-1$ cases of the same object category, and then evaluate on the held-out sequence.
As shown in Table~\ref{tab:zero-shot}, our method outperforms all baselines by a clear margin across geometry, tracking, and rendering metrics.

Figure~\ref{fig:qual_unseen} further provides qualitative comparisons under unseen interaction and unseen object settings.
For unseen interactions, our method follows the observed deformation trajectory more faithfully, while PhysTwin shows larger deviations as the interaction pattern differs from those used for training.
For unseen objects, \modelname still maintains stable geometry and produces plausible deformation.
\paragraph{Cross-scene consistency.}
We further compare the cross-scene variance of estimated physical parameters.
For each object category, we measure how the predicted spring-mass parameters vary across different scenes and interactions.
As shown in Fig.~\ref{fig:cross-scene statics}, our method yields much lower parameter variance than the baseline, indicating that it assigns more consistent physical properties to similar objects.
By contrast, the baseline exhibits larger case-to-case fluctuations, suggesting that per-scene inverse fitting may over-adapt parameters to individual sequences.
This demonstrates that our material-aware prediction leads to more stable and transferable physical estimates.
\subsection{Ablation study}
Table~\ref{tab:ablation} reports the ablation results of our main components.
Removing the learned codebook leads to the largest performance drop, confirming that the codebook provides a more stable and structured physical parameter space than direct regression.
Removing the material regularizer $\mathcal{L}_{prior}$ also degrades performance, particularly for future prediction, indicating that semantic material knowledge helps regularize physical parameter learning under weak visual supervision.
Finally, without part decomposition, the model predicts physical parameters at the object level instead of the part level, which weakens its ability to model spatially heterogeneous material properties.

\section{Conclusion}
We presented MatPhys, a framework for reconstructing simulatable digital twins of deformable objects from single-view interaction videos. By combining part-level material decomposition, VLM-guided material priors, and codebook-based parameter prediction, MatPhys recovers geometry, appearance, and physical parameters for resimulation and future prediction. Experiments show competitive reconstruction performance and stronger generalization to future frames and unseen interactions.
These results suggest that material-aware feed-forward parameter prediction is a promising step toward editable and simulation-ready deformable digital twins.

Our method is still limited by the dataset scale, category diversity, and relatively smooth interaction patterns. Future work will expand the dataset with more diverse objects, materials, and interactions to improve real-world generalization.

{\small
\bibliographystyle{ieeenat_fullname}
\bibliography{sample-base}
}

\end{document}